\documentclass[a4paper,fleqn]{cas-dc}

\usepackage{natbib}
\usepackage{subfigure}
\usepackage{hyperref}       
\usepackage{url}            
\usepackage{amsfonts}       
\usepackage{nicefrac}       
\usepackage{microtype}      
\usepackage{xcolor}         
\usepackage{amssymb}
\usepackage{graphicx}
\usepackage{tabularx}
\usepackage{multirow}
\usepackage{paralist}
\usepackage{enumitem}
\usepackage{booktabs}
\usepackage{caption}
\usepackage[switch]{lineno}

\newcommand{\ie}{\textit{i.e.}\xspace}
\newcommand{\eg}{\textit{e.g.}\xspace}
\newcommand{\etal}{\textit{et al.}\xspace}

\newcommand{\gt}{\textit{gt}~}
\newcommand{\posneg}{\textit{pos/neg}~}
\newcommand{\pos}{\textit{pos}~}

\def\tsc#1{\csdef{#1}{\textsc{\lowercase{#1}}\xspace}}
\tsc{WGM}
\tsc{QE}
\tsc{EP}
\tsc{PMS}
\tsc{BEC}
\tsc{DE}

\begin{document} 
\begin{sloppypar}
\let\WriteBookmarks\relax
\def\floatpagepagefraction{1}
\def\textpagefraction{.001}
\shorttitle{Detecting tiny objects in aerial images: A normalized Wasserstein distance and a new benchmark}
\shortauthors{C. Xu, et~al.}

\title [mode = title]{Detecting tiny objects in aerial images: A normalized Wasserstein distance and a new benchmark}                      

\author[1]{Chang Xu}[style=chinese]
                        
\cormark[1]

\ead{xuchangeis@whu.edu.cn}

\address[1]{School of Electronic Information, Wuhan University, Wuhan 430072, China}

\author[1]{Jinwang Wang}[style=chinese]
\cormark[1]
\ead{jwwangchn@whu.edu.cn}

\author[1]{Wen Yang}[style=chinese]
\cormark[2]
\ead{yangwen@whu.edu.cn}


\author[1]{Huai Yu}[style=chinese]
\ead{yuhuai@whu.edu.cn}

\author[1]{Lei Yu}[style=chinese]
\ead{ly.wd@whu.edu.cn}

\author[2]{Gui-Song Xia}[style=chinese]
\ead{guisong.xia@whu.edu.cn}
\address[2]{School of Computer Science, Wuhan University, Wuhan 430072, China}

\cortext[cor1]{Equal contribution}
\cortext[cor2]{Corresponding author}

\begin{abstract}
Tiny object detection (TOD) in aerial images is challenging since a tiny object only contains a few pixels. State-of-the-art object detectors do not provide satisfactory results on tiny objects due to the lack of supervision from discriminative features. Our key observation is that the Intersection over Union (IoU) metric and its extensions are very sensitive to the location deviation of the tiny objects, which drastically deteriorates the quality of label assignment when used in anchor-based detectors. To tackle this problem, we propose a new evaluation metric dubbed Normalized Wasserstein Distance (NWD) and a new RanKing-based Assigning (RKA) strategy for tiny object detection. The proposed NWD-RKA strategy can be easily embedded into all kinds of anchor-based detectors to replace the standard IoU threshold-based one, significantly improving label assignment and providing sufficient supervision information for network training. Tested on four datasets, NWD-RKA can consistently improve tiny object detection performance by a large margin.
Besides, observing prominent noisy labels in the Tiny Object Detection in Aerial Images (AI-TOD) dataset, we are motivated to meticulously relabel it and release AI-TOD-v2 and its corresponding benchmark. In AI-TOD-v2, the missing annotation and location error problems are considerably mitigated, facilitating more reliable training and validation processes. Embedding NWD-RKA into DetectoRS, the detection performance achieves 4.3 AP points improvement over state-of-the-art competitors on AI-TOD-v2. Datasets, codes, and more visualizations are available at: \url{https://chasel-tsui.github.io/AI-TOD-v2/}.

\end{abstract}


\maketitle

\section{Introduction}

Tiny objects are ubiquitous in aerial images, meanwhile detecting tiny objects in aerial images has many application scenarios, including vehicle detection, traffic condition monitoring, and maritime rescue. Even though object detection has achieved significant progress thanks to the development of deep neural networks~\citep{Faster-R-CNN_2015_NIPS, Focal-Loss_2017_ICCV, FCOS_2019_ICCV}, most of them are dedicated to detecting objects of normal size. While tiny objects (less than $16\times16$ pixels in the AI-TOD dataset~\citep{AI-TOD_2020_ICPR}) in aerial images often exhibit extremely limited appearance information, which poses great challenges in learning discriminative features, leading to enormous failure cases when detecting tiny objects~\citep{SNIPER_2018_NIPS, AI-TOD_2020_ICPR, TinyPerson_2020_WACV}.    

Recent advances for tiny object detection (TOD) mainly focus on improving the feature discrimination ability~\citep{FPN_2017_CVPR,M2Det_2019_AAAI,DetectoRS_2020_CVPR,PGAN_2017_CVPR,SOD-MTGAN_2018_ECCV,Better_to_Follow_2019_ICCV}. 
Some efforts have been devoted to normalizing the input image scales to enhance the resolution of small objects and corresponding features~\citep{SNIP_2018_CVPR,SNIPER_2018_NIPS}. While the Generative Adversarial Network (GAN) is proposed to directly generate super-resolved representations for small objects~\citep{PGAN_2017_CVPR,SOD-MTGAN_2018_ECCV,Better_to_Follow_2019_ICCV}. 
Besides, the Feature Pyramid Network (FPN) is proposed to learn multi-scale features to achieve scale-invariant detectors~\citep{FPN_2017_CVPR,M2Det_2019_AAAI,DetectoRS_2020_CVPR}. Indeed, existing approaches have improved TOD performance to some extent, but the precision boost is commonly achieved with additional cost.

\begin{figure*}[t]
    \centering
    \subfigure[Tiny scale object (vehicle)]{
    \label{fig:tiny_analysis1}
    \includegraphics[width=0.48\linewidth]{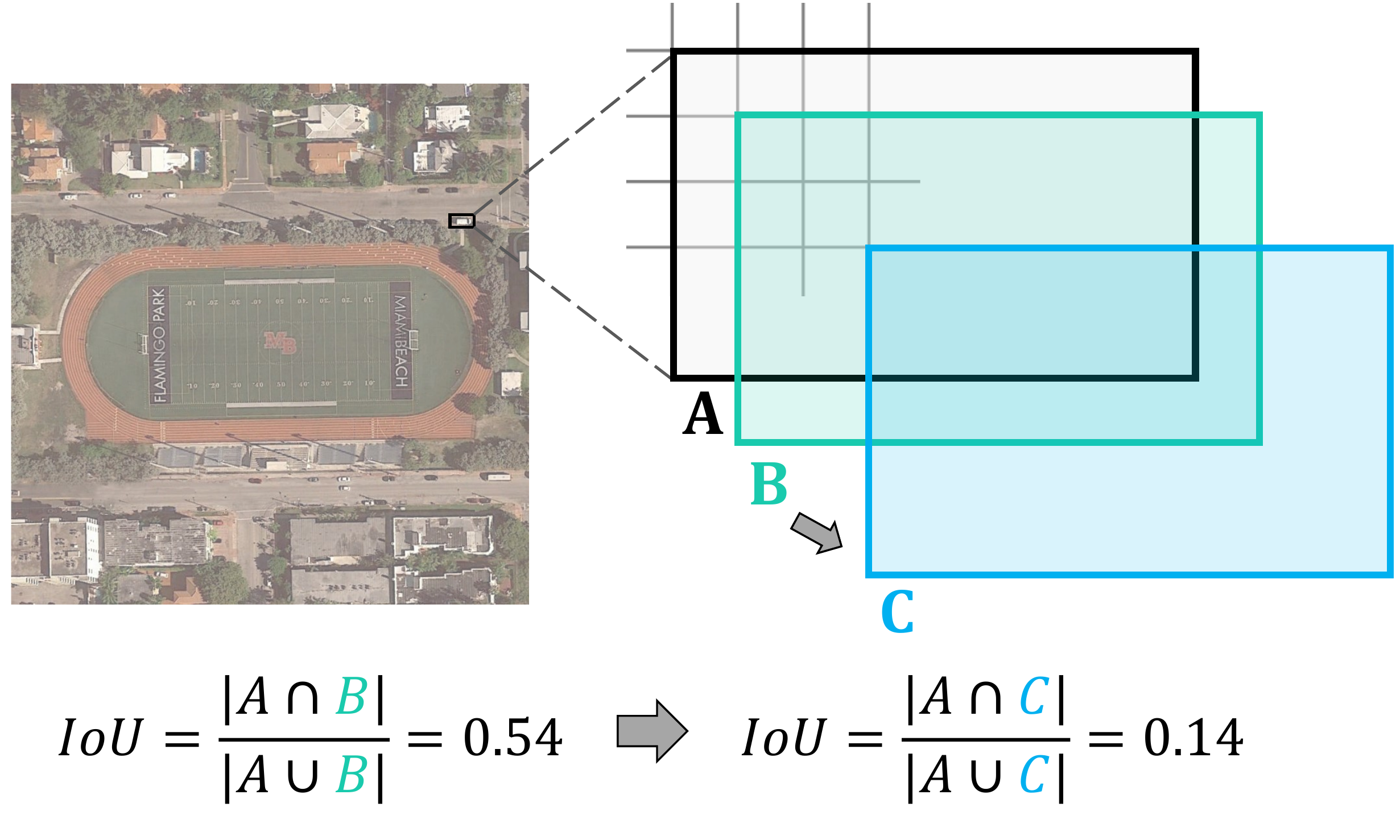}
    }
    \subfigure[Normal scale object (playground)]{
    \label{fig:tiny_analysis2}
    \includegraphics[width=0.48\linewidth]{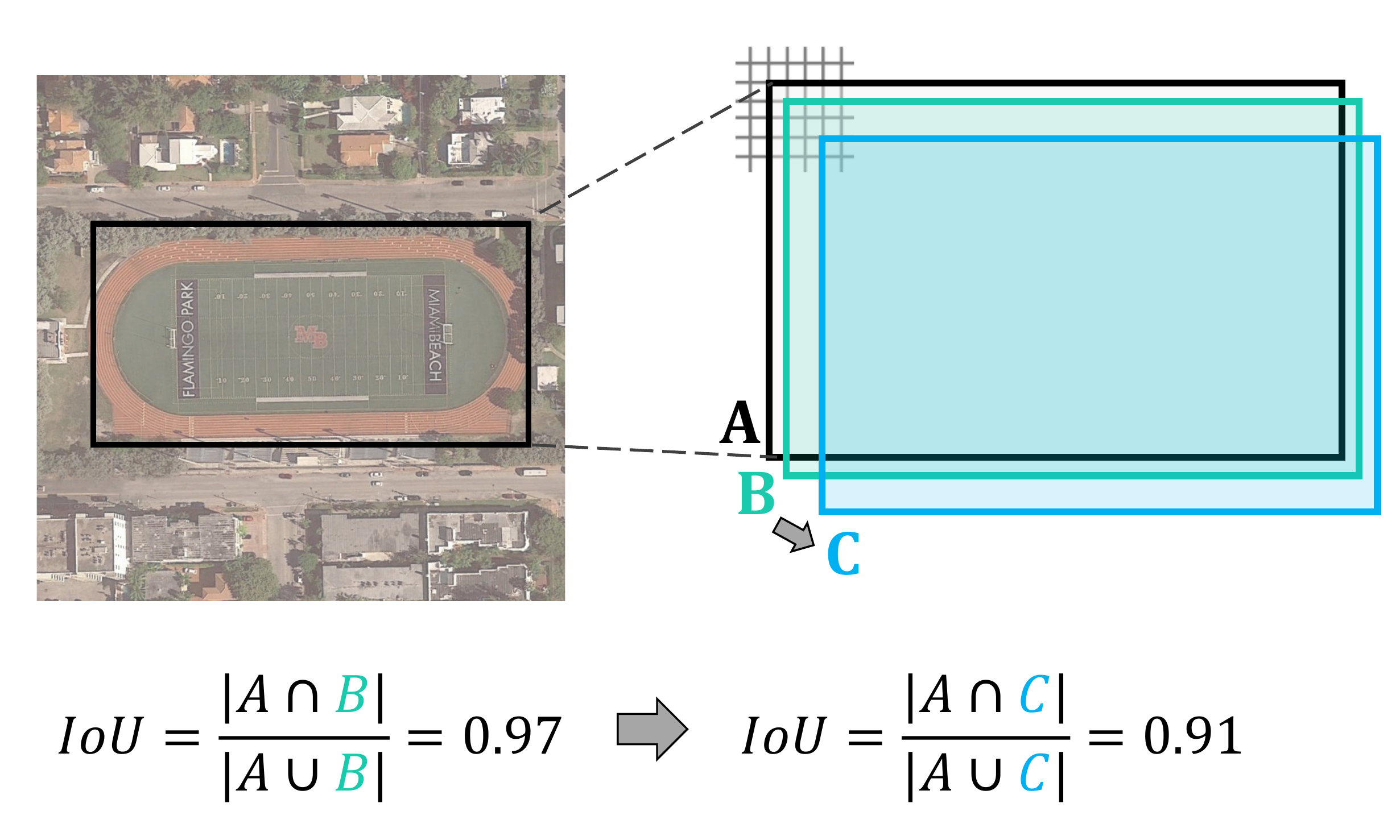}
    }
    \caption{The sensitivity analysis of IoU on tiny and normal scale objects. Note that each grid denotes a pixel, box $A$ denotes the ground truth bounding box, box $B$, $C$ denote the predicted bounding box with 1 pixel and 3 pixels right-down deviation, respectively.}
    \label{fig:tiny_analysis}
\end{figure*}

\begin{figure*}[t]
    \centering
    \subfigure
    {
        \label{fig:offset1}
        \includegraphics[width=0.22\linewidth]{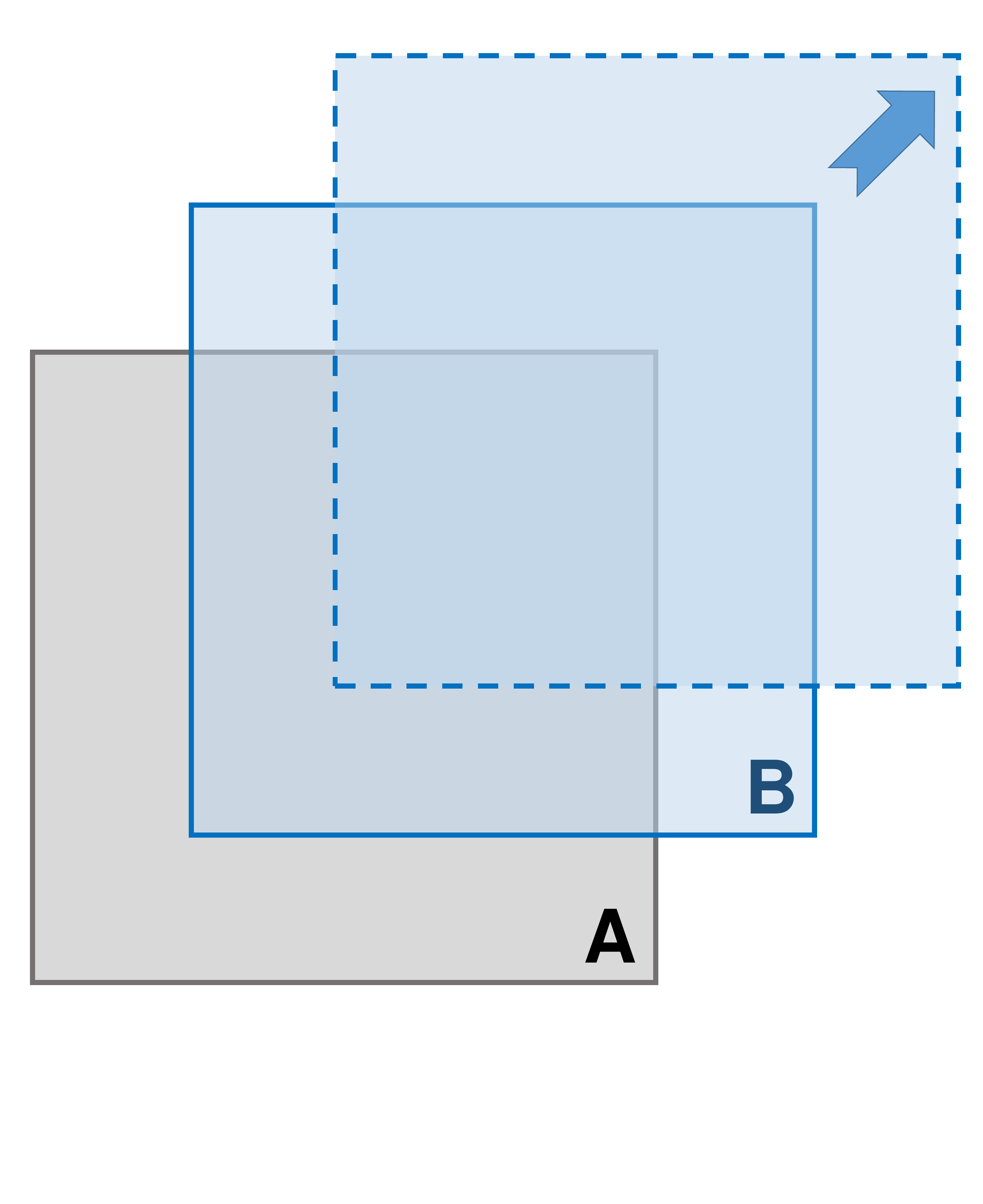}
    }
    \subfigure
    {
        \label{fig:iou1}
        \includegraphics[width=0.36\linewidth]{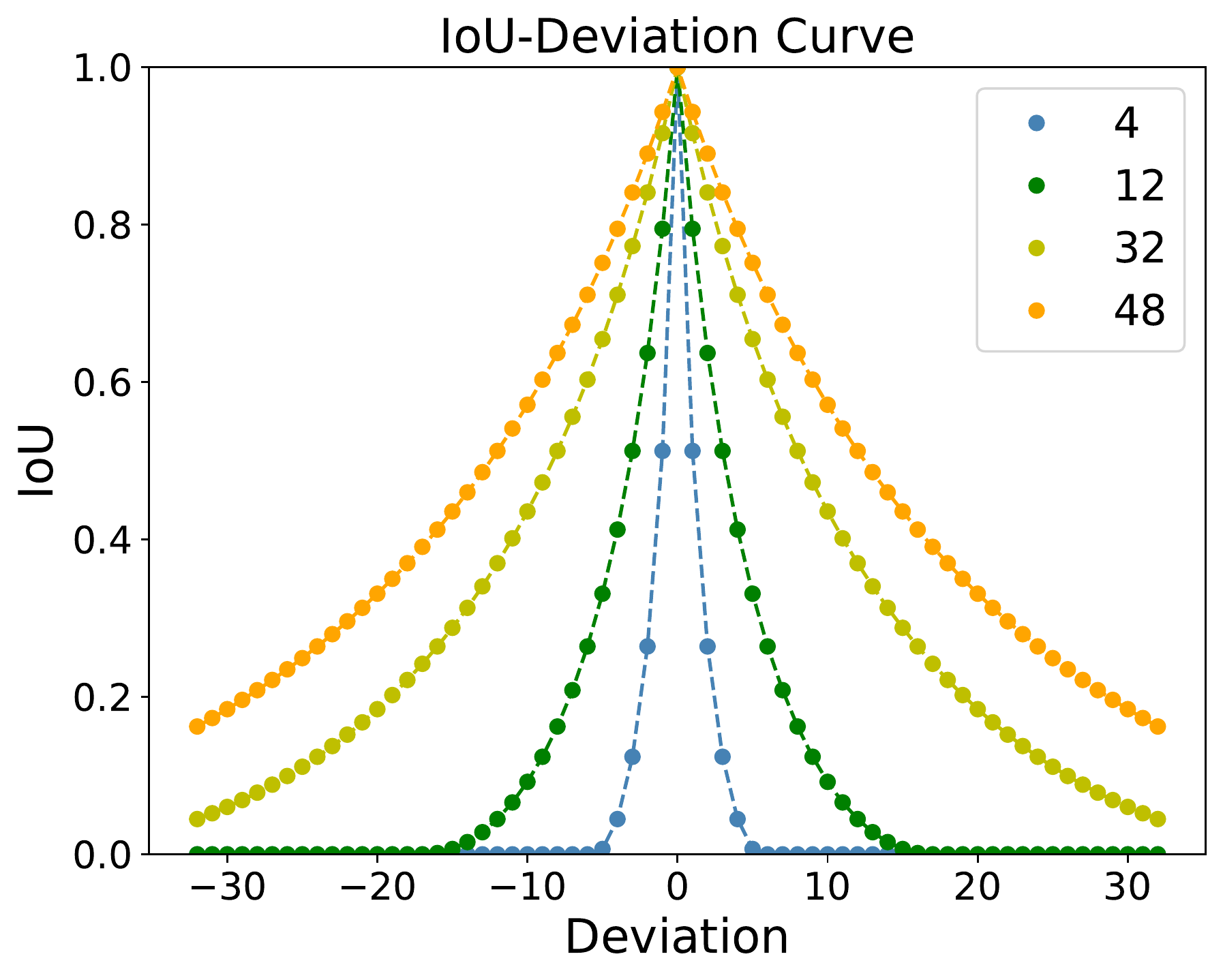}
    }
    \subfigure
    {
        \label{fig:nwd1}
        \includegraphics[width=0.36\linewidth]{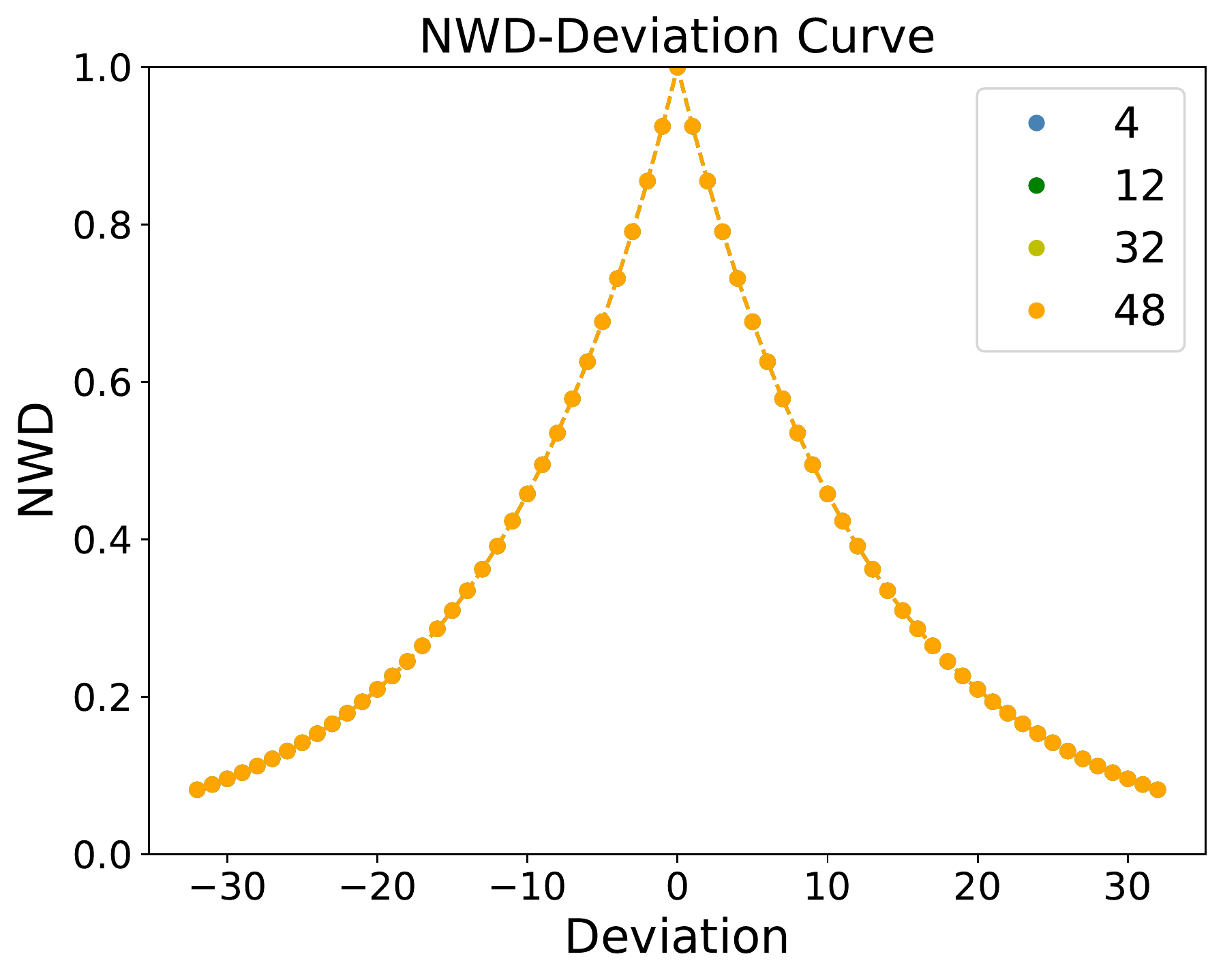}
    }
    
    \vspace{-3mm}
    \subfigure
    {
        \label{fig:offset2}
        \includegraphics[width=0.22\linewidth]{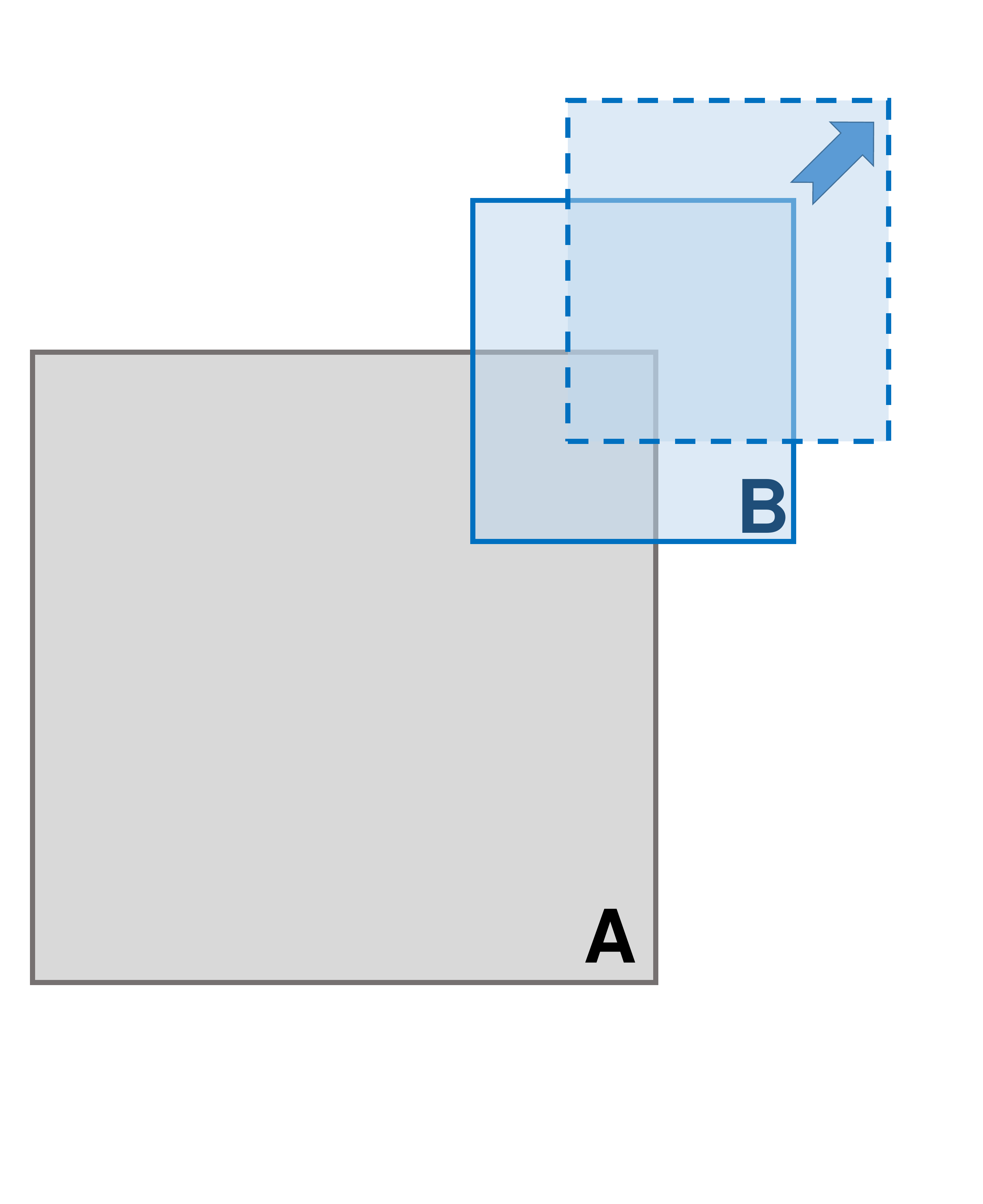}
    }
    \subfigure
    {
        \label{fig:iou2}
        \includegraphics[width=0.36\linewidth]{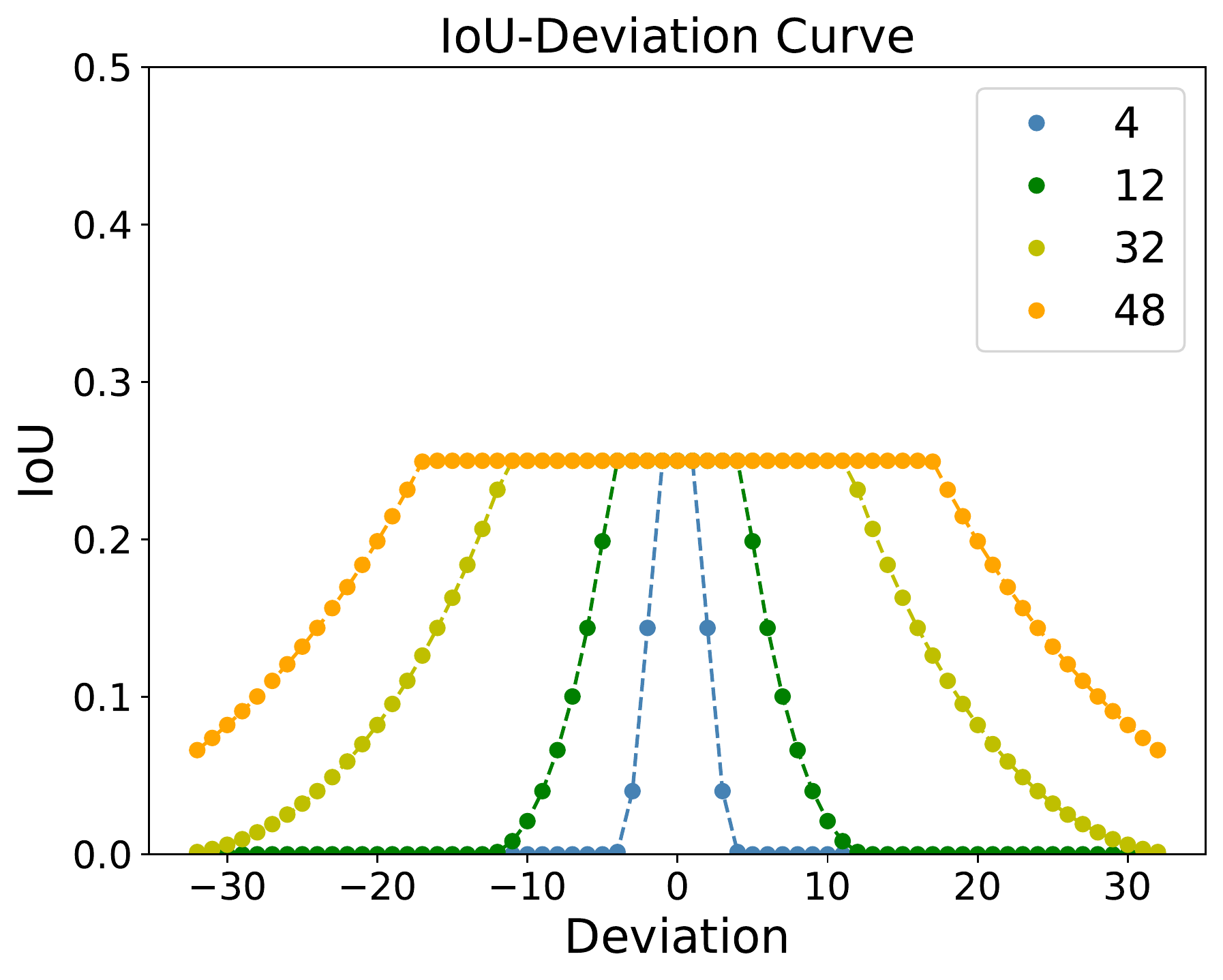}
    }
    \subfigure
    {
        \label{fig:nwd2}
        \includegraphics[width=0.36\linewidth]{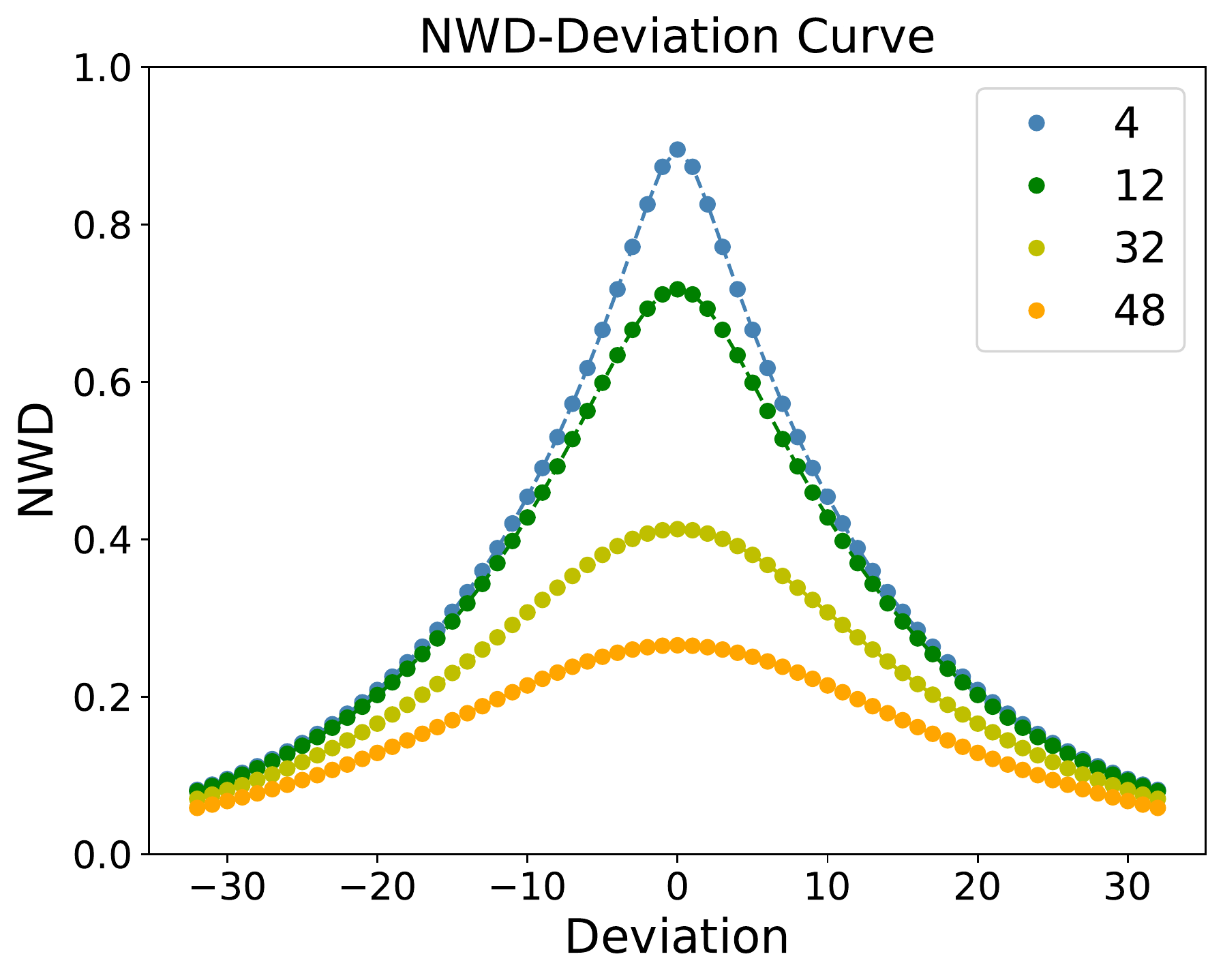}
    }
    \vspace{-3mm}
    \caption{A comparison between the IoU-Deviation Curve and the NWD-Deviation Curve in two different scenarios. The legend with numbers denotes the scale setting of the box $A$. In the first row, the scale of the box $B$ is set the same as box $A$, while in the second row, the side length of the box $B$ is set to the half of box $A$. Curves of different colors refer to the IoU-Deviation or NWD-Deviation curves under the corresponding scale setting. The abscissa value denotes the number of pixels deviation between the center points of $A$ and $B$, and the ordinate value denotes the corresponding metric value. Note that the location of the bounding box can only change discretely. Therefore, the Value-Deviation curves are presented in the form of scatter diagrams.} 
    \label{fig::deviation_analysis}
\end{figure*}

In addition to learning discriminative features, the quality of the training sample selection plays a significant role in anchor-based tiny object detectors \citep{ATSS_CVPR_2020} where the assignment of positive and negative (\textit{pos/neg}) labels is essential. However, for a tiny object, the properties of only a few pixels will increase the difficulty of training sample selection. As shown in Fig.~\ref{fig:tiny_analysis} and Fig.~\ref{fig::deviation_analysis}, we have the following two key observations. \textbf{The most striking observation} is that the sensitivity of IoU to objects with different scales is of large variance. Specifically, for the tiny object with $5\times 8$ pixels, a minor location deviation will lead to a notable IoU drop (from $0.54$ to $0.14$), resulting in inaccurate label assignment. However, for the normal object with $200\times 320$ pixels, the IoU changes slightly (from $0.97$ to $0.91$) with the same location deviation. Moreover, Fig.~\ref{fig::deviation_analysis} shows four IoU-Deviation curves with different object scales, and the curve declines faster as the object size becomes smaller. It is worth noting that the sensitivity of IoU results from the particularity that the location of the bounding box can only change discretely. In other words, the value change of IoU from unit pixel deviation gradually approximates a continuous change as the object scale increases, so that the discreteness of IoU is usually ignored when detecting normal-sized objects. 
\textbf{The second observation} is that IoU fails to reflect the positional relationship of two bounding boxes when they have no overlap or are mutually inclusive (\ie, the value of IoU keeps constant), which is often the case for tiny bounding boxes. 

The above observations imply that IoU is unsuitable for evaluating the positional relationship between tiny objects and easily leads to the following three drawbacks in label assignment. Specifically, IoU thresholds ($\theta_p$, $\theta_n$) are used to assign \textit{pos/neg} training samples in anchor-based detectors, for example, ($0.7$, $0.3$) are used in Region Proposal Network (RPN)~\citep{Fast-R-CNN_2015_ICCV}. Firstly, the sensitivity of IoU on a tiny object makes a minor location deviation flip the anchor label. Most anchor candidates become \textit{neg} training samples under current way of measurement (\ie, IoU), leading to tiny objects' imbalanced number of \textit{pos/neg} training samples.  Secondly, we find that IoU-based assignment suffers from severe scale imbalance problem. In the AI-TOD dataset~\citep{AI-TOD_2020_ICPR}, the average number of positive samples assigned to each ground-truth (\textit{gt}) is extremely imbalanced according to their scales, where objects of larger scales tend to be assigned with more \textit{pos} samples than objects of very tiny scales. Therefore, the network is skewed towards optimizing objects of larger scales. Although dynamic assignment strategies such as ATSS~\citep{ATSS_CVPR_2020} can adaptively set IoU thresholds for assigning \textit{pos/neg} labels according to the statistical characteristics of objects, the sensitivity of IoU makes it difficult to find a good threshold and provide high-quality \textit{pos/neg} samples for tiny object detectors. Finally, the sample compensation assignment strategy does not work well under the current way of measurement. To make each instance sufficiently trained, heuristics are usually deployed to guarantee at least one training sample for each \gt~\citep{Faster-R-CNN_2015_NIPS, s3fd_2017_cvpr, dotd_2021_cvprw}, we call this method sample compensation strategy. Nevertheless, IoU keeps zero when two boxes are separated. In this case, it is hard to provide a rational rank of anchor candidates according to the same IoU score.

\begin{figure}[t]
    \centering
    \subfigure{
    \label{fig:noise1}
    \includegraphics[width=0.45\linewidth]{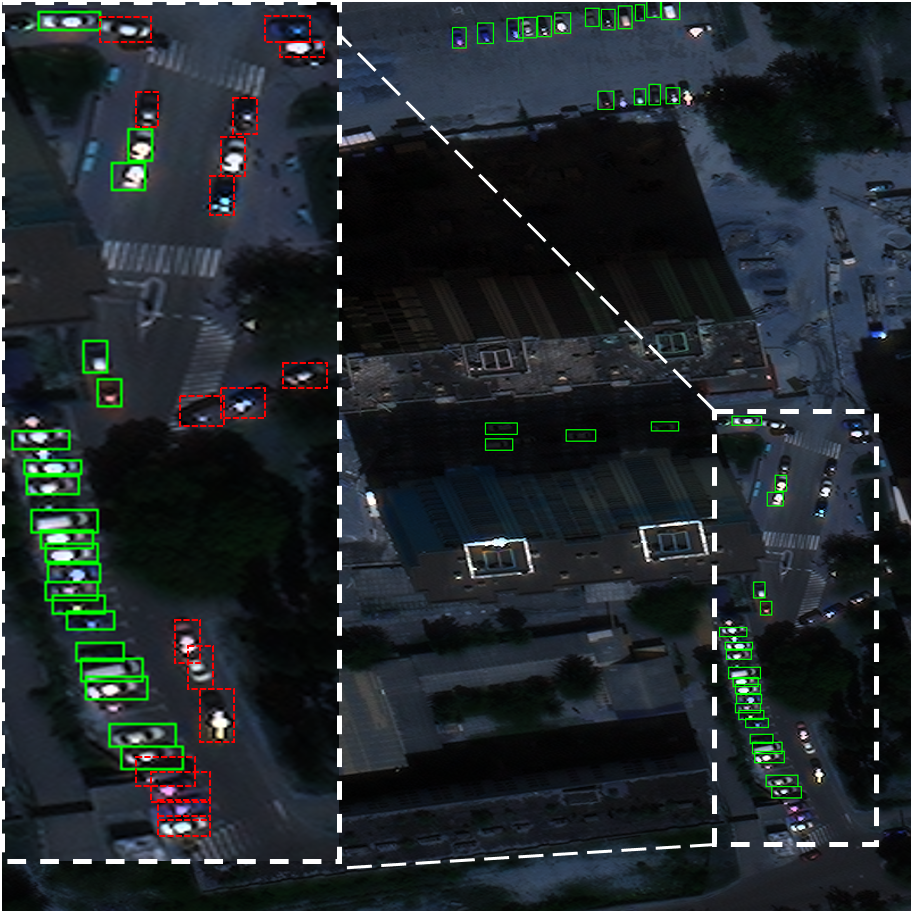}
    }
    \subfigure{
    \label{fig:noise2}
    \includegraphics[width=0.45\linewidth]{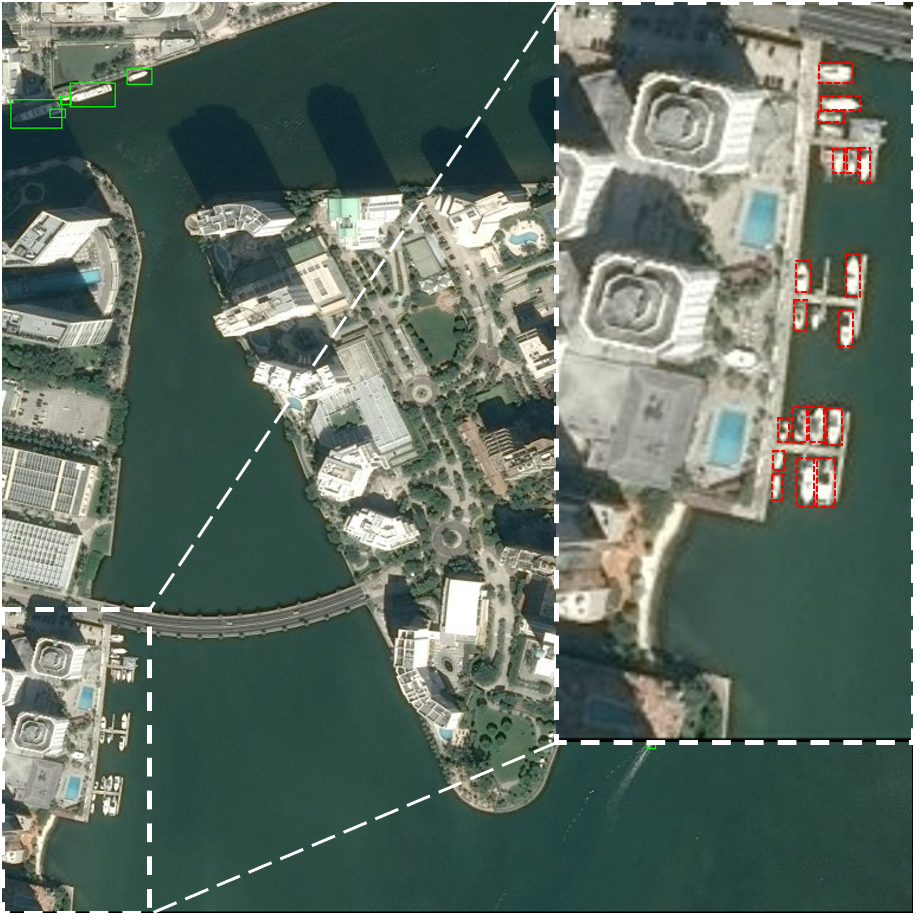}
    }
    \caption{Visualization of typical noisy labels in AI-TOD dataset. Green boxes denote the original labels, red dotted boxes denote RoIs missed to be annotated. }
    \label{fig:label_noise}
\end{figure}

On the other hand, high-quality data plays a pivotal role in deep learning methods since current deep learning networks highly demand data quality. However, the dataset's quality for tiny object detection in aerial images is far from satisfactory. Tiny objects usually exhibit extremely few pixels, so they are easy to be confused with complex and diverse backgrounds, especially in aerial images~\citep{AI-TOD_2020_ICPR}, which increases the difficulty and workload of labelling tiny objects. To date, very little attention has been paid to the detailed annotation or quality check of a dataset specialized for tiny objects in aerial images. Existing tiny object detection benchmarks include TinyPerson benchmark~\citep{TinyPerson_2020_WACV} and AI-TOD benchmark~\citep{AI-TOD_2020_ICPR}. Unfortunately, the TinyPerson benchmark is established for single-class person detection in realistic scenarios. In the first tiny object detection dataset in aerial images AI-TOD~\citep{AI-TOD_2020_ICPR}, we observe that there exists much label noise, as shown in Fig.~\ref{fig:label_noise}, in which the problem of objects missed to be annotated is pronounced.
Thus, improving the annotation quality and establishing a more reliable benchmark for tiny object detection in aerial images is necessary.

To address the problems mentioned above, we attempt to push forward the development of TOD in aerial images from the perspectives of methodologies and datasets. Specifically, we seek to design a unified way that can be applied to existing anchor-based detectors and boost their performance when detecting tiny objects in aerial images. Based on the problems of IoU when assigning training samples, in this paper, we propose a new metric to measure the similarity of tiny bounding boxes and combine it with a customized label assignment strategy. Concretely, we first model the bounding boxes as 2-D Gaussian distributions, then use the classic Wasserstein Distance to measure the similarity of derived Gaussian distributions. Furthermore, the exponential nonlinear transformation function is selected to remap the Gaussian Wasserstein distance and normalize its value range, and we call the obtained metric Normalized Wasserstein Distance (NWD). For tiny objects, it is often the case that their bounding boxes have no overlap with almost all anchors. The major advantage of Wasserstein distance is that it can measure the distribution similarity even if the overlap of distributions is negligible. Hence, the proposed NWD can consistently reflect the positional relationship between non-overlapping tiny objects, which the IoU metric cannot realize.
In addition, the NWD metric is smooth to location deviations and has the characteristic of \textit{scale balance}. As shown in Fig.~\ref{fig::deviation_analysis}, the NWD-Deviation curves of different scales coincide, which means that more positive samples can be assigned to tiny objects when directly switching IoU to NWD under the threshold-based assigning strategy.

To sufficiently leverage the potential of our proposed NWD metric, we propose combining it with a customized RanKing-based training sample Assignment (RKA) strategy and further improving tiny object detection performance in aerial images. Different from the threshold-based assignment strategy in Faster R-CNN~\citep{Faster-R-CNN_2015_NIPS}, we seek to rank anchor candidates according to their NWD scores with ground truth boxes, warranting that each instance can be assigned with sufficient positive training samples. 
NWD-RKA can considerably alleviate the three issues arising from IoU-threshold-based one (\ie, \posneg imbalance, scale imbalance, and failure of sample compensation), thereby consistently improving TOD performance.

In addition, we propose to convene experts to relabel our preliminary TOD dataset~\citep{AI-TOD_2020_ICPR} meticulously. According to our statistics, the label noise problem is significantly alleviated, and more than 50 thousand  previously missed instances to be annotated are increased. The released AI-TOD-v2 contains eight categories with 28,036 images and 752,754 instances. Moreover, the mean absolute object size of AI-TOD-v2 is only 12.7 pixels, which, as far as we know, is the dataset of the smallest object size among all object detection datasets. Then, to facilitate algorithm developments and comparisons with AI-TOD-v2, we establish a benchmark that contains more baselines than the preliminary version~\citep{AI-TOD_2020_ICPR}.

The contributions of this paper are three-fold: 

\begin{itemize}
    \item We propose NWD-RKA as a better training sample assignment strategy for tiny objects, which can simultaneously alleviate the three drawbacks of IoU threshold-based strategy (\ie, severe \posneg sample imbalance for tiny objects, scale-sample imbalance, and failure of sample-compensation). 
    \item We carefully relabel AI-TOD and release AI-TOD-v2, where the label noise problem is significantly alleviated. Besides, we establish a comprehensive benchmark by several baseline detectors. The training/validation images and annotations are publicly available.
    \item Our proposed NWD-RKA can be applied to all kinds of anchor-based detectors and boost their performance on tiny objects. On the AI-TOD-v2 dataset, we achieve the performance of 24.7 AP and 57.2 $\rm{AP_{0.5}}$, which outperforms the state-of-the-art competitors by a large margin. Moreover, significant improvements can be seen on AI-TOD, VisDrone2019, and DOTA-v2.0 datasets.    
\end{itemize}

The rest of this paper is organized as follows. In Section 2, we briefly describe the related works. In Section 3, we show the details of AI-TOD-v2. A detailed description of the proposed method, including the modeling of NWD and the design of NWD-RKA-based detectors, is then provided in Section 4. Next, an in-depth analysis of different metrics and NWD-RKA is given in Section 5. Then, the experimental results are discussed in Section 6. Finally, we conclude this paper in Section 7.

\section{Related Work}
\subsection{Aerial Object Detection Datasets}
 In aerial object detection, many datasets have been proposed to promote its development. For example, DIOR~\citep{DIOR_2019_ISPRS}, DOTA~\citep{DOTA_2018_CVPR}, xView~\citep{xview_2018_arXiv},  VisDrone~\citep{VisDrone-2018-Det_2018_ECCV},  HRSC2016~\citep{HRSC2016_2016}, VEDAI~\citep{VEDAI_2016_JVCIR}, NWPU VHR-10~\citep{NWPU_VHR-10_2014_ISPRSJ}, UAVDT~\citep{UAVDT_2019_IJCV}, and FAIR1M~\citep{fair1m_2021_arxiv} are all public aerial object detection datasets. Nevertheless,  these datasets' average absolute object sizes are much larger than 32 pixels, and objects larger than 32 pixels occupy most of the datasets, which indicates that these aerial image datasets are not suitable for evaluating the performance of tiny object detectors. Although there exist some datasets focusing on tiny object detection tasks (\eg, TinyPerson~\citep{TinyPerson_2020_WACV},  $\mathcal{R}^2\mbox{-}\rm{CNN}$~\citep{R2CNN_2019_TGRS}, TinyPerson only contains a single class of person in realistic scenarios, and the dataset in $\mathcal{R}^2\mbox{-}\rm{CNN}$ is not publicly available. 
 
 In comparison, our proposed AI-TOD-v2 is dedicated to tiny object detection in aerial images. The mean absolute object size of AI-TOD-v2 is only 12.7 pixels, and about 86\% instances are smaller than 16 pixels, which is far smaller than existing datasets. 
 
\subsection{Tiny Object Detection Strategies}
Most of the previous small/tiny object detection strategies can be roughly grouped into the following five categories: multi-scale feature learning, context-based detection, data augmentation, designing better training strategy~\citep{reviewtod_2020_icv} and label assignment strategy.

\textbf{Multi-scale Feature Learning:} A simple and classic way is to resize input images into different scales and train different detectors, each of which can attain the best performance in a certain range of scales. It improves the detection performance of tiny objects with additional computation costs. To reduce computation cost, some works~\citep{SSD_2016_ECCV,ffssd_2018_ICGIP, MultiScaleCNN_2018_ISPRSJ,  FPN_2017_CVPR, RefineDet_2018_CVPR, R-DFPN_2018_RS, hynet_2021_isprs} try to construct feature maps of different scales. For instance, SSD~\citep{SSD_2016_ECCV} detects objects from feature layers of different resolutions, and Feature Pyramid Network (FPN)~\citep{FPN_2017_CVPR} constructs a top-down structure with lateral connections and combines feature information of different scales, improving tiny object detection performance. Based on these two fundamental networks, many works, including Feature Fused SSD (FFSSD)~\citep{ffssd_2018_ICGIP}, Multi-Scale CNN (MSCNN)~\citep{MultiScaleCNN_2018_ISPRSJ} and Deep Feature Pyramid Network (DFPN)~\citep{R-DFPN_2018_RS} are presented.

\textbf{Context-based Detection:} Objects are closely related to the background and environment information in the image. Context information plays an important role in object detection, especially when an object is tiny and its feature information is limited. Multi-Region CNN (MRCNN)~\citep{mrcnn_2015_cvpr} utilizes local context information by extracting features from sub-regions of object proposals and simply concatenating these features. Inside-Outside Network (ION)~\citep{Inside_Outside_Net_2016_CVPR} exploits global context information by using features both inside and outside the Region of Interests (RoI). Relation Network~\citep{relationnet_2018_cvpr} proposed by Hu \etal establishes an association model between objects by the interaction between appearance feature and geometry. 

\textbf{Data Augmentation:} It is universally acknowledged that performance improvement of detectors can be achieved by using more data for training. Identically, the performance of tiny object detectors can be improved by data augmentation. A simple yet effective way is to collect more tiny object data and make high-quality datasets. Some other simple data augmentation approaches include image flipping, up-sampling, down-sampling, rotating, etc. Krisantal \etal~\citep{augmentation_2019_arxiv} find that one of the factors for the deterioration of tiny object detection is the lack of small object representation in datasets. And Krisantal  \etal~\citep{augmentation_2019_arxiv} propose to augment data by oversampling images containing tiny objects and copy-pasting tiny objects. 

\textbf{Designing Better Training Strategy:} Inspired by the observation that it is challenging to detect tiny objects and large objects simultaneously, Singh~\etal~propose SNIP~\citep{SNIP_2018_CVPR} and SNIPER~\citep{SNIPER_2018_NIPS} to selectively train objects within a particular scale range. Besides, Kim~\etal~\citep{san_2018_eccv} introduce Scale-Aware Network (SAN) and map the features extracted from different spaces onto a scale-invariant subspace, making detectors more robust to scale variation.

\begin{figure*}
    \centering
    \includegraphics[width=0.98\linewidth]{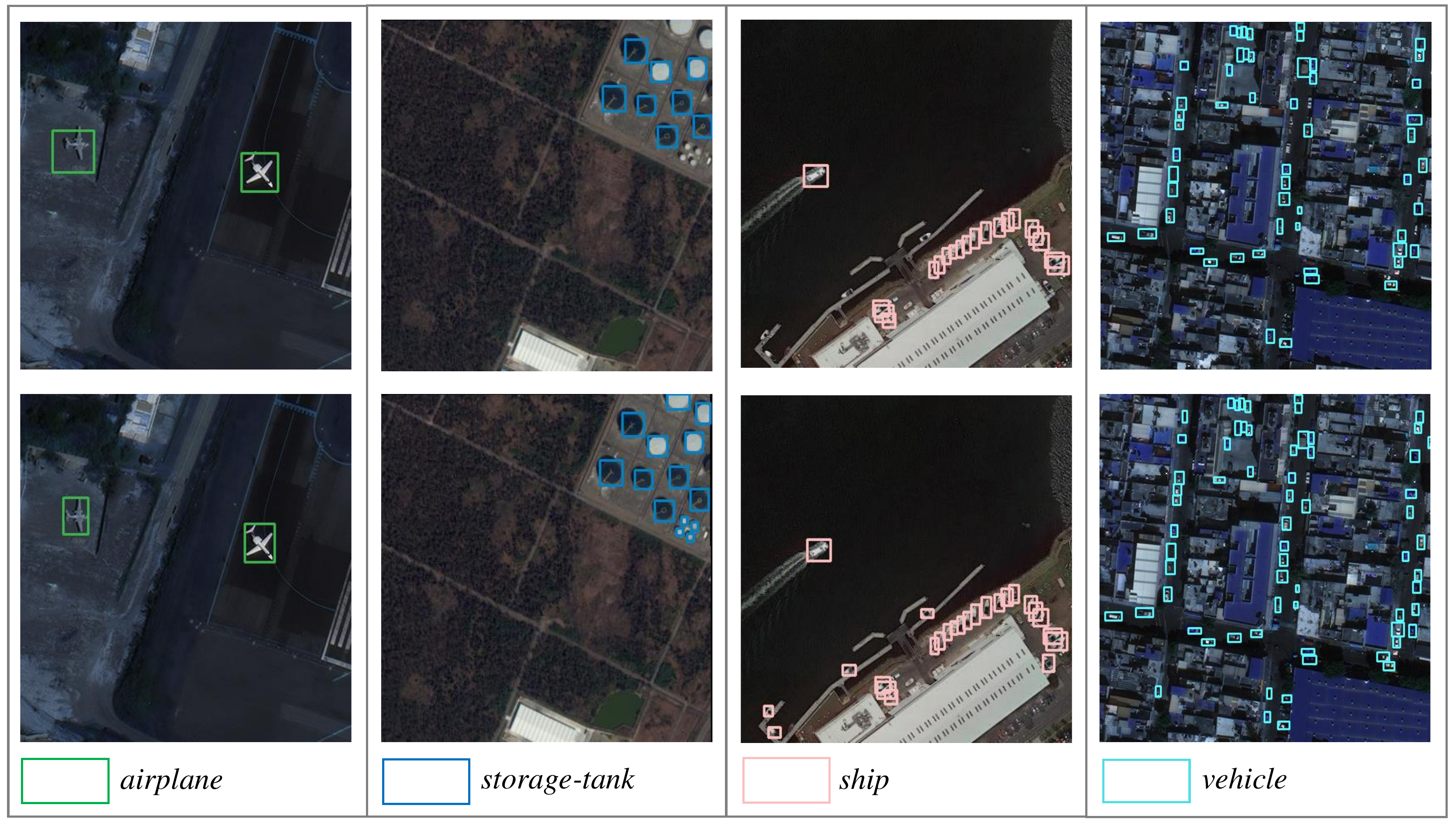}
    \caption{A comparison between AI-TOD and AI-TOD-v2. We choose four out of eight categories to illustrate the improvement of annotations. Note that images in the first row are from AI-TOD, and images in the second row are from AI-TOD-v2.}
    \label{fig:dataset_comparison}
\end{figure*}

\textbf{Label Assignment Strategy:} It is challenging to assign high-quality anchors to \gt boxes of tiny objects. A simple way is to lower the IoU threshold when selecting positive samples. Although it can make tiny objects match more anchors, the overall quality of training samples will deteriorate. Besides, many recent works try to make the label assignment process more adaptive, aiming at improving the detection performance~\citep{ota_2021_cvpr}. For instance, Zhang \etal~\citep{ATSS_CVPR_2020} propose an Adaptive
Training Sample Selection (ATSS) to automatically compute the \posneg threshold for each \gt by statistic value of IoU from a set of anchors. Probabilistic Anchor Assignment (PAA)~\citep{PAA_2020_ECCV} assumes that the distribution of joint loss for \posneg samples follows the Gaussian distribution. In addition, Optimal Transport Assignment (OTA)~\citep{ota_2021_cvpr} formulates the label assignment process as an Optimal Transport problem from a global perspective. However, these methods all use the IoU metric to measure the similarity between two bounding boxes and mainly focus on the label assignment's threshold setting, which is not suitable for TOD. 

\textbf{Super-resolution-based Strategies:} As revealed in the previous work~\citep{srsatellite_2019_cvprw}, increasing the resolution of images by deep super-resolution frameworks can improve object detection performance in satellite images. Some methods propose integrating Super-Resolution (SR) strategies into the detection pipeline to enhance the feature representation and boost the detection performance of tiny objects. For instance, an edge-enhanced SR GAN~\citep{edgegan_2020_rs} is proposed to enhance the remote sensing images, thereby improving the detection performance of small objects. By adding an auxiliary GAN into the detection framework, the work~\citep{auxiliarygan_2021_rs} enhances the quality of SR architecture, increasing the object detection performance. Moreover, the cyclic GAN and residual feature aggregation are incorporated to improve the SR framework~\citep{bashir2021small}, significantly improving the detection performance of small objects.

\textbf{Core Detection Adaptation Strategies:} Some studies seek to adapt the core detection to work with tiny objects directly. Mask-guided SSD~\citep{maskguidedssd_2020} proposes to enhance detection features with contextual information and eliminate background features with a segmentation mask, showing good performance on small object detection. YOLO-Fine~\citep{yolofine_2020_rs} improves the YOLO to adapt to the small object detection task, and it simultaneously holds a high efficiency and accuracy on small object detection. In addition, a modified Faster R-CNN~\citep{ren2018small} is designed for detecting small objects by increasing the resolution of FPN and combining several training strategies.

In contrast, our method mainly focuses on designing a better similarity measurement metric and its customized label assignment strategy, which can be used to replace IoU-based label assignment strategies in object detectors. 

\section{Dataset}
We build the AI-TOD-v2 dataset based on the preliminary work AI-TOD~\citep{AI-TOD_2020_ICPR}. We observe that there exist many objects missed to be annotated, which mainly results from the fact that AI-TOD is constructed based on publicly available aerial image datasets~\citep{DOTA_2018_CVPR,xview_2018_arXiv,VisDrone-2018-Det_2018_ECCV,Airbus-ship_2018_kaggle,DIOR_2019_ISPRS} which are not dedicated to tiny object detection. The label noise problem significantly affects the training and validation of tiny object detectors. This motivates us to meticulously relabel the AI-TOD dataset so that the training and validation of detectors are more reliable. Moreover, we establish a more comprehensive benchmark with more baseline detectors to encourage the research of tiny object detection in aerial images. Note that AI-TOD and AI-TOD-v2 share the same images but contain different annotations.

\subsection{Dataset Optimization}

The original AI-TOD dataset includes 28,036 images with 700,621 objects of 8 classes annotated with Horizontal Bounding Box (HBB)~\citep{AI-TOD_2020_ICPR}. Herein, we keep the images and classes fixed, also we keep the representation form of HBB since tiny objects only contain a few pixels and most of them can be roughly viewed as points. We ask annotators to meticulously adjust the annotations. The detailed process is described as follows. 

To start with, we train our proposed tiny object detector, which holds the state-of-the-art performance on the AI-TOD dataset, and visualize its detection results. Specifically, we first respectively train our detector on the {\tt trainval set} and {\tt test set}, we then use the trained model to respectively generate detection results on the corresponding {\tt trainval set} and {\tt test set}. Finally, we visualize the detection bounding box by dividing them into True Positive (TP), False Negative (FN), and False Positive (FP) detections with a confidence threshold of 0.3. According to our observation, some FP detections are strong indications of objects that were previously missed to be labelled, which can serve as hints for relabelling.

With the assistance of the visualization results, we carefully select some typical images and convene experts in the field of aerial image interpretation to meticulously relabel them. Based on the annotation and discussion of experts, we establish an annotation manual and train the volunteers for the large-scale annotation. The volunteers are recruited from Wuhan University who have a background in aerial image interpretation and computer vision. 

Then, the large-scale relabelling process can be separated into two stages. In the first stage, the volunteers adjust the noisy labels that they are highly confident with and simultaneously mark down the names of images containing objects or annotations that they are not confident about. In the second stage, we work in teams to decide the precise location of the objects and use a voting strategy to determine the exact category of ambiguous objects in the image marked down by volunteers.

To guarantee the quality of annotations, we conduct a double-blind check. We randomly exchange the annotations of different annotators and encourage them to find the annotations that they think are of low quality. Finally, we collect these low-quality annotations and use the teamwork mentioned above and voting strategy to determine the precise location and category of objects. Some comparisons of annotations before and after relabelling are visualized as in Fig.~\ref{fig:dataset_comparison}. We can observe that many missed objects are supplemented, and location errors are fixed.

\begin{table}
	\centering
	\caption{Number of objects per image set and per class. The number in the left side of each column denotes the instance number in AI-TOD, the number in the right side of each column denotes the instance number in AI-TOD-v2.}
	\resizebox{\linewidth}{!}{
	\begin{tabular}{lllll}  
	\toprule
	v1/v2  & Train & Validation & Trainval & Test \\
	 \midrule
    AI         & 623/666  & 170/177 & 793/843 & 745/855\\
    BR    & 512/603  & 140/152 & 652/755 & 689/803\\
    ST  & 5269/5584 & 2477/2635 & 7746/8219 & 5860/6090\\
    SH     & 13539/14074 & 3791/3839 & 17330/17913 & 17633/18946  \\
    SP      & 293/787  &  34/112 & 327/899 & 292/466 \\
    VE    & 248042/265194  &  59904/64247  & 307964/319441 & 306665/332604\\
    PE   & 14126/14446 & 3841/3860 & 17967/18306 & 15443/16030 \\
    WM   & 176/180 & 67/69 & 243/249 & 290/335 \\
    \midrule
   Total       &  282580/301534 & 70424/75091 & 353004/376625 & 347641/376129  \\
	 
	\bottomrule
	\end{tabular}}
	\label{tab:instance_number}
\end{table}

\subsection{Characteristics of AI-TOD-v2 Dataset}
The AI-TOD-v2 is the dataset with the smallest object size in the earth observation community. We compare the statistical characteristics of AI-TOD-v2 with our previous AI-TOD and some other datasets from the following four aspects.

\textbf{Instance number.} AI-TOD-v2 contains totally 28,036 aerial images with 752,745 annotated object instances. Compared with the number of 700,621 in the first version, the increased 52,133 instances further suggest the significance of precise relabeling of existing TOD datasets. For each object category and each image set (\ie, {\tt train set}, {\tt validation set}, {\tt trainval set}, {\tt test set}), the thorough number of object instances is reported in Tab.~\ref{tab:instance_number}. Images and annotations of the {\tt train set} and {\tt validation set} are publicly available.    

\begin{figure*}[t]
    \centering
    \subfigure[]{
    \label{fig:class_numb}
    \includegraphics[width=0.23\linewidth]{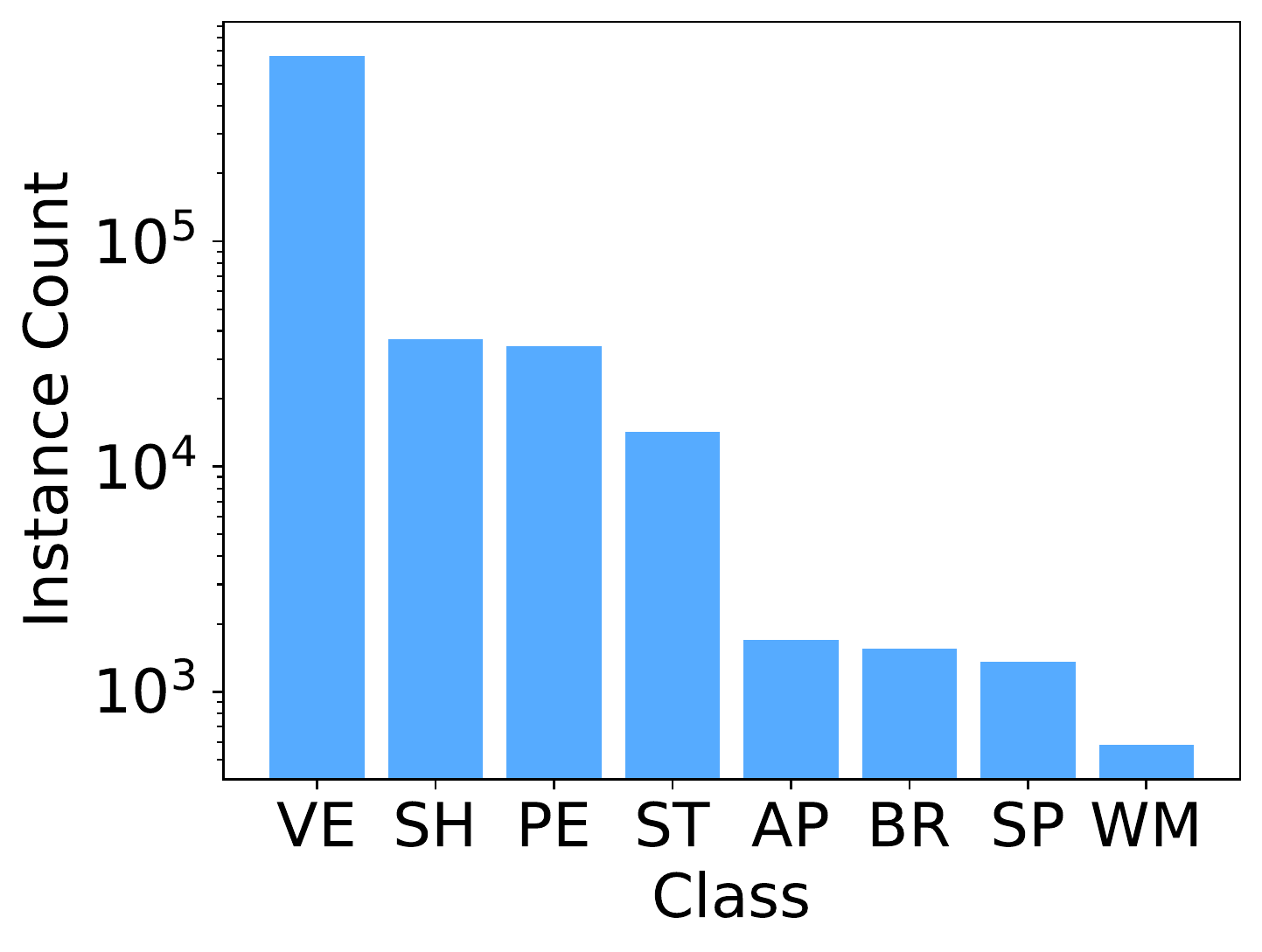}
    }
    \subfigure[]{
    \label{fig:image_instance_num}
    \includegraphics[width=0.23\linewidth]{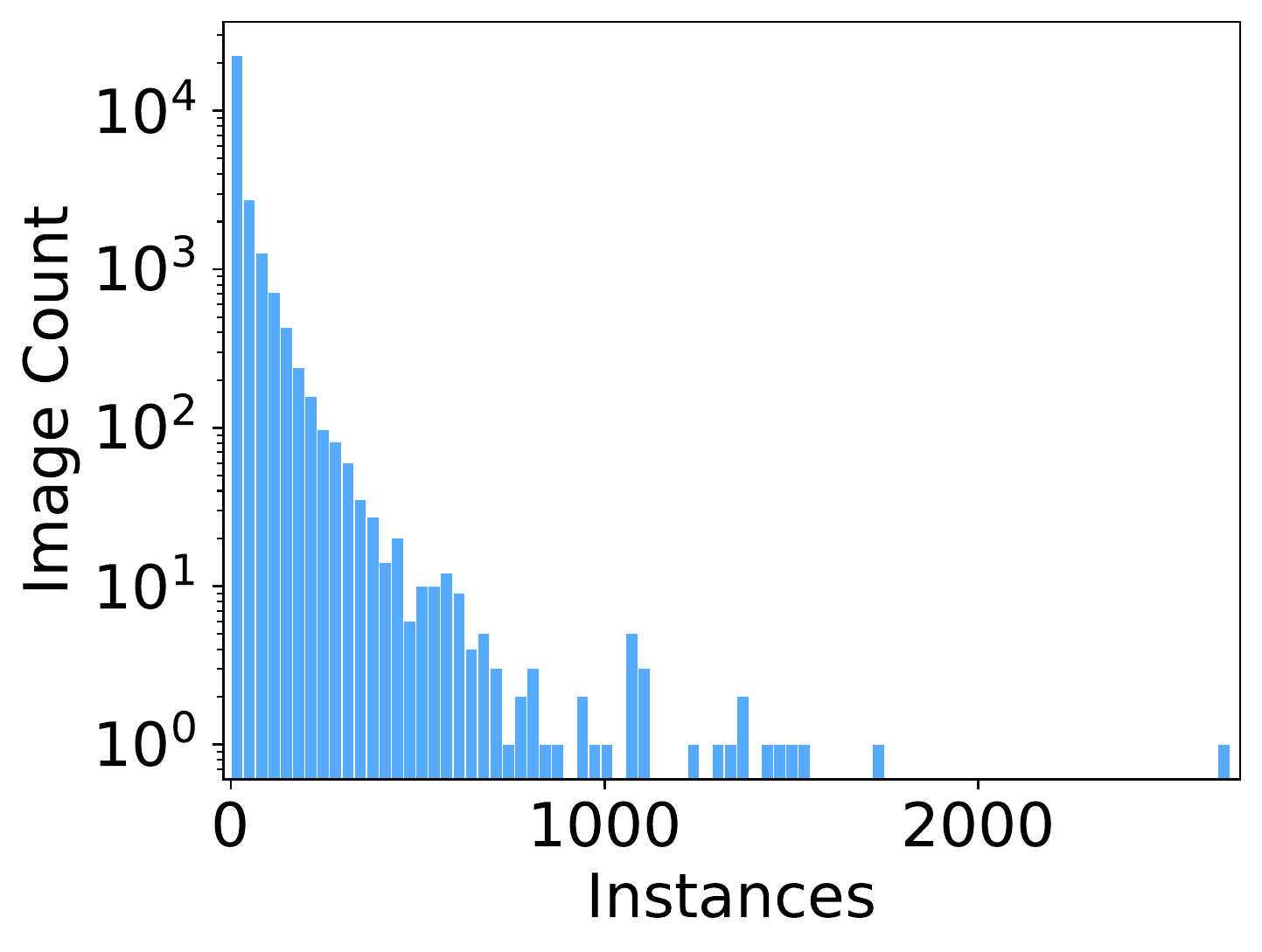}
    }
    \subfigure[]{
    \label{fig:size_distribution}
    \includegraphics[width=0.23\linewidth]{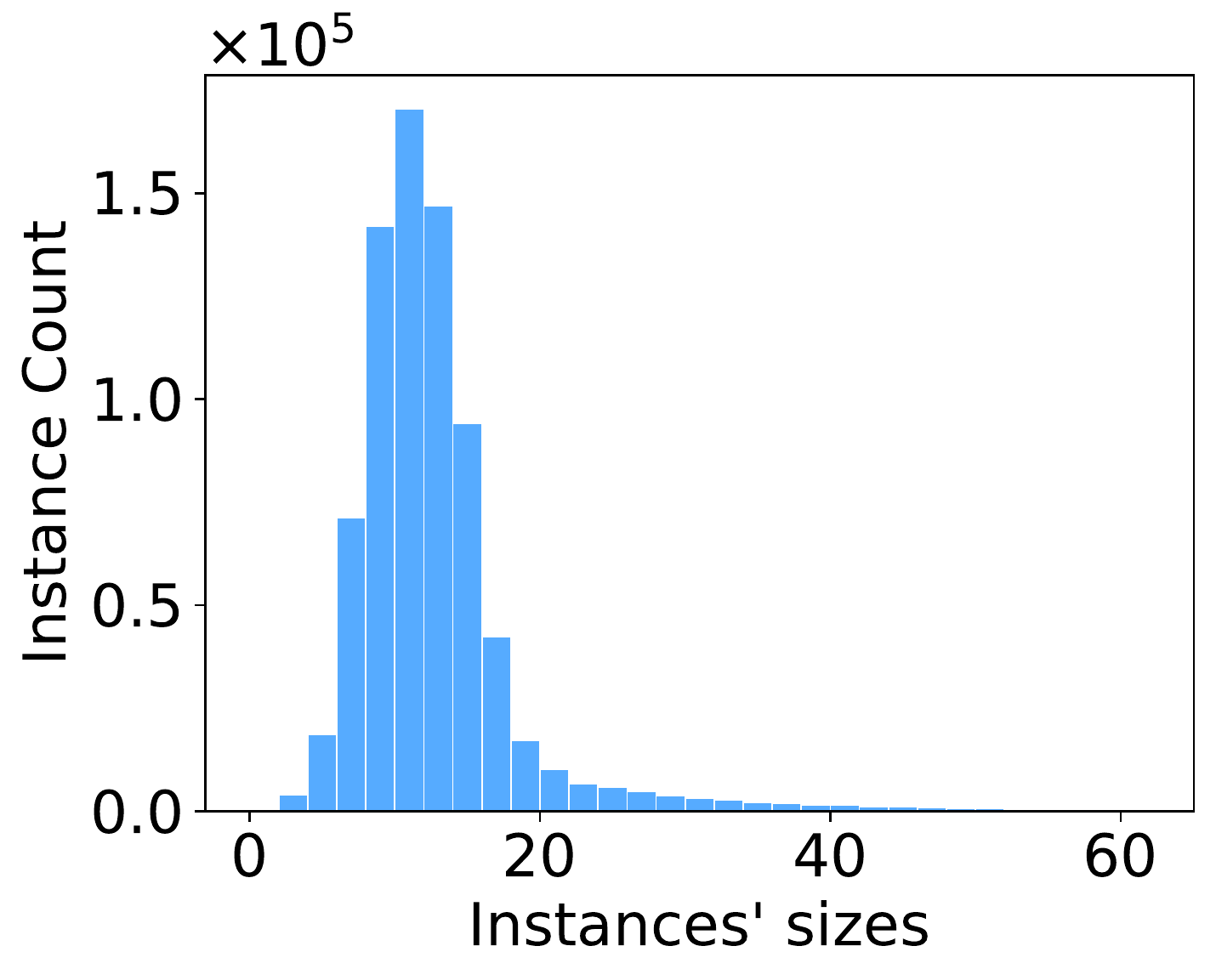}
    }
    \subfigure[]{
    \label{fig:class_size}
    \includegraphics[width=0.23\linewidth]{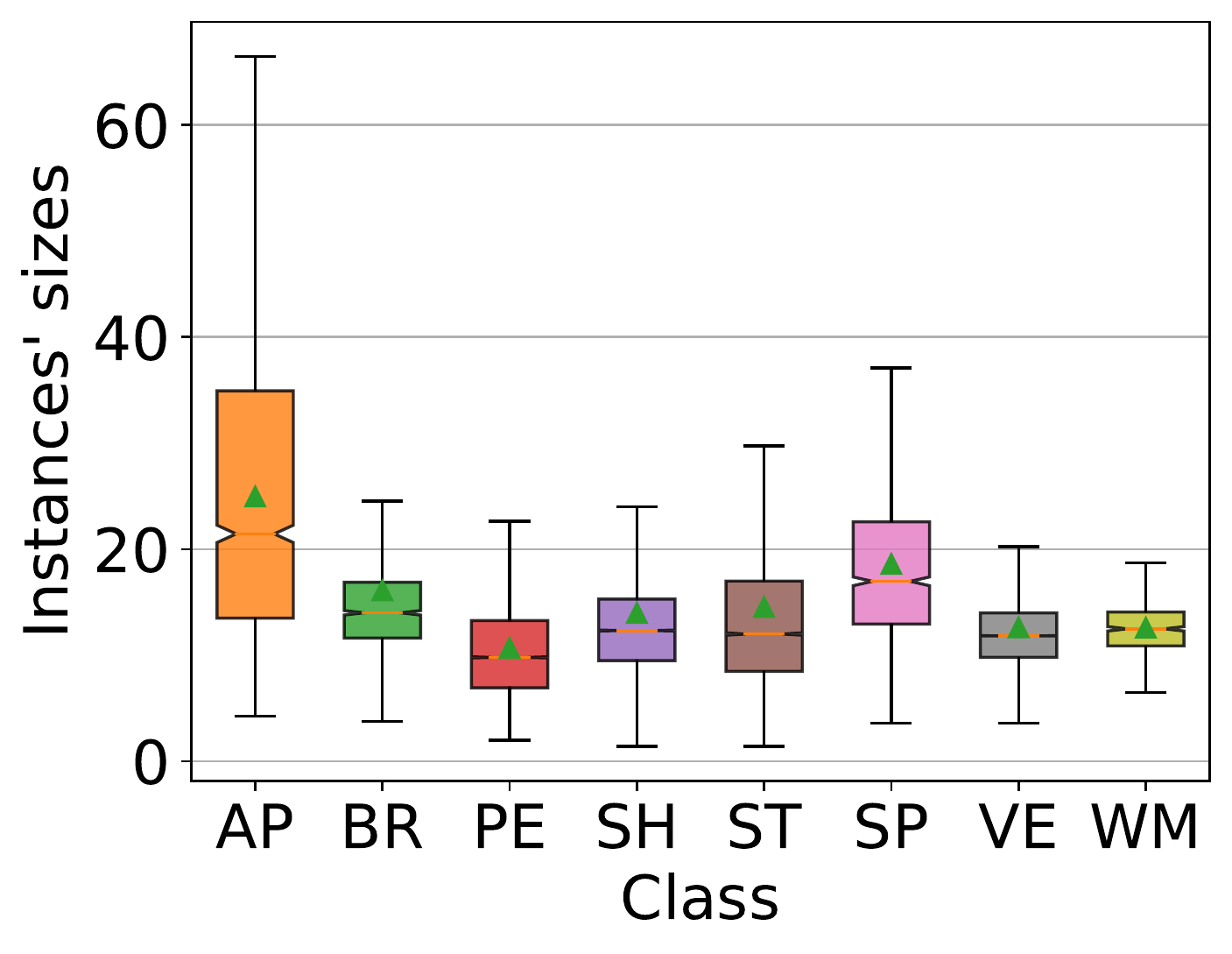}
    }
    \caption{Statistics of instances in AI-TOD-v2. (a) Histogram of the instance number per class. (b) Histogram of instance number per image. (c) Histogram of number of instances’ sizes. (d) Boxplot depicting the range of sizes for each object category.}
    \label{fig:aitodv2_stat}
\end{figure*}

\textbf{Object count per category.} Same with the previous version, there are totally 8 categories of common-seen tiny objects in AI-TOD-v2, including: $airplane\; \rm{(AI)}$,\; $bridge\; \rm{(BR)}$,\; $storage\mbox{-}tank\; \rm{(ST)}$,\; $ship\; \rm{(SH)}$,\; $swimming\mbox{-}pool\; \rm{(SP)}$,\; $vehicle\; \rm{(VE)}$,\; $person\; \rm{(PE)}$, and $wind\mbox{-}mill\; \rm{(WM)}$. Fig.~\ref{fig:class_numb} shows the histogram of the number of instances each class. From the histogram, an imbalanced number of different classes can be observed (\ie, the quantity of vehicles is much larger than that of wind-mill), such a class imbalance is consistent with real life scenarios and brings much challenge to the design of algorithms.

\textbf{Object size distribution.} The size distribution of AI-TOD-v2 is shown in Fig.~\ref{fig:size_distribution}. It is apparent from the histogram that the absolute object size is mainly distributed around 12 pixels. A comparison of the mean and standard deviation of the absolute object size between AI-TOD-v2 and other aerial object detection images is shown in Tab.~\ref{tab:scales}. The absolute object size of AI-TOD-v2 is only 12.7 pixels, which is marginally smaller than AI-TOD and much smaller than other datasets.
 
\textbf{Object size of different categories.} The size range of different categories in AI-TOD is shown in Fig.~\ref{fig:class_size}. In order to analyze the scales of tiny objects more specifically, we consider objects with absolute size in the range [2,8] as $very~tiny$, [8,16] as $tiny$, [16,32] as $small$ and [32,64] as $medium$. The percentages of $very~tiny$, $tiny$, $small$ and $medium$ objects in AI-TOD-v2 is 12.4\%, 73.4\%, 12.4\% and 1.8\%, respectively. And it can be seen that most categories of objects are only in the range of $very~tiny$ and $tiny$.

\begin{table}
	\centering
	\caption{Mean and standard deviation of object scale on different datasets.}
	\resizebox{\linewidth}{!}{
	\begin{tabular}{ccc}  
	\toprule
	Dataset  & Absolute Size (pixels) & Relative Size (pixels) \\
	 \midrule
    DIOR         & $65.7\pm 91.8$  &  $0.082\pm 0.115$  \\
    DOTA-v1.0    & $55.3\pm 63.1$  &  $0.028\pm 0.034$  \\
    Airbus-Ship  & $44.9\pm 44.1$  &  $0.058\pm 0.057$  \\
    VisDrone     & $35.8\pm 32.8$  &  $0.030\pm 0.026$  \\
    xView        & $34.9\pm 39.9$  &  $0.011\pm 0.013$  \\
    DOTA-v1.5    & $34.0\pm 47.8$  &  $0.016\pm 0.026$  \\
    VEDAI(512)   & $33.4\pm 11.3$  &  $0.065\pm 0.022$  \\
    AI-TOD    & $12.8\pm 5.9$  &  $0.016\pm 0.007$  \\
    \midrule
    AI-TOD-v2       & $12.7\pm 5.6$   &  $0.016\pm 0.007$  \\
	\bottomrule
	\end{tabular}}
	\label{tab:scales}
\end{table}

\begin{table}[h]
	\centering
	\caption{The similarity check between the results generated by NWD-RKA from AI-TOD ($\rm{\textbf{NA}}_{1}$) and the annotations of AI-TOD-v2 ($\rm{\textbf{A}}_{2}$). Note that in the first experiment, the NWD-RKA is self-trained and self-validated on the {\tt trainval set} and {\tt test set} of AI-TOD, while in the second experiment, we calculate the AP between $\rm{\textbf{NA}}_{1}$ and $\rm{\textbf{A}}_{2}$.}
	\resizebox{\linewidth}{!}{
	\begin{tabular}{lcccccccc}  
	\toprule
	 Set & AP & $\rm{AP_{0.5}}$ & $\rm{AP_{0.75}}$ & $\rm{AP_{vt}}$ & $\rm{AP_{t}}$ & $\rm{AP_{s}}$ & $\rm{AP_{m}}$ \\
	 \midrule
	AI-TOD {\tt trainval} & 32.5 & 68.8 & 26.4 & 11.2  & 32.7 & 38.1 & 50.4 \\
    AI-TOD {\tt test} & 34.0 & 70.5 & 28.0 & 14.7  & 34.7 & 41.3 & 50.4 \\
	\midrule
	Cross {\tt trainval} & \textcolor[RGB]{0,150,0}{32.4} & \textcolor{red}{69.8} & \textcolor[RGB]{0,150,0}{25.6} & 11.2 & \textcolor[RGB]{0,150,0}{32.4} & \textcolor[RGB]{0,150,0}{37.9} & \textcolor{red}{51.4}  \\
	Cross {\tt test} & \textcolor[RGB]{0,150,0}{33.5} & \textcolor{red}{70.9} & \textcolor[RGB]{0,150,0}{26.8} & \textcolor[RGB]{0,150,0}{14.1}  & \textcolor[RGB]{0,150,0}{33.9} & \textcolor[RGB]{0,150,0}{39.6} & \textcolor{red}{51.6} \\
	\bottomrule
	\end{tabular}}
	\label{tab:similarity_check}
\end{table}
\begin{figure}[h]
        \centering
        \includegraphics[width=0.99\linewidth]{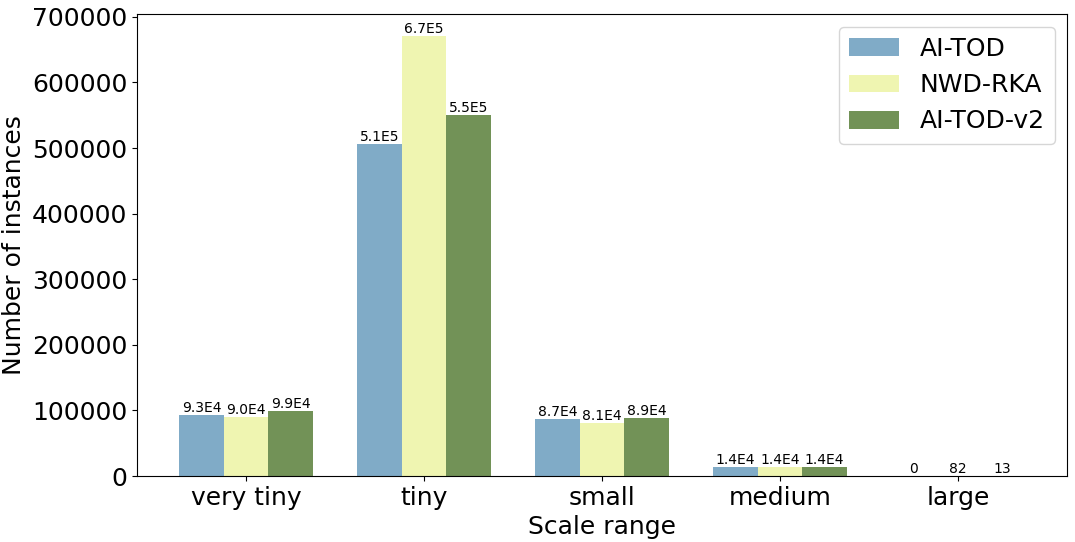}
        \caption{A comparison of instance number in different scale ranges. The NWD-RKA denotes the predictions generated on AI-TOD by the proposed NWD-RKA with DetectoRS.}
        \label{fig:compare_scale_distribution}
\end{figure}
 
\subsection{Evaluation of AI-TOD-v2 Dataset}
To objectively evaluate the annotations of the proposed AI-TOD-v2, we perform a similarity check and some statistical analyses. 
In the similarity check, we mainly compare the similarity between bounding boxes generated with NWD-RKA and the re-annotated bounding boxes by AP metric. There are two groups of experiments in this similarity check. In the first group, the NWD-RKA is self-trained and self-validated on the AI-TOD {\tt trainval set} and {\tt test set} respectively. In the second group, we crossly verify the similarity between AI-TOD-v2 and the predictions generated by NWD-RKA from AI-TOD. Concretely, we treat the annotations of AI-TOD-v2 as \textit{ground truth} and treat the predictions generated by NWD-RKA from AI-TOD as \textit{predictions to be evaluated}. For simplicity, we mark the annotations of AI-TOD as $\rm{\textbf{A}}_{1}$, the predictions generated by NWD-RKA from AI-TOD as $\rm{\textbf{NA}}_{1}$, and the annotations of AI-TOD-v2 as $\rm{\textbf{A}}_{2}$ in the following statement. The results are listed in  Tab.~\ref{tab:similarity_check}. It can be observed that the AP metric between $\rm{\textbf{NA}}_{1}$ and $\rm{\textbf{A}}_{2}$ is only around $33.0$ AP, especially, in the $very~tiny$ scale, the metric is lower than $15.0$ $\rm{AP}_{vt}$. Moreover, we compare the AP metric between the first group (\textit{i.e.,}~$\rm{\textbf{NA}}_{1}~vs.~\rm{\textbf{A}}_{1}$) and the second group (\textit{i.e.,}~$\rm{\textbf{NA}}_{1}~vs.~\rm{\textbf{A}}_{2}$). The AP metric of the second group is slightly lower than the that of the first group, indicating that the similarity of $\rm{\textbf{NA}}_{1}~vs.~\rm{\textbf{A}}_{2}$ is lower than the similarity of  $\rm{\textbf{NA}}_{1}~vs.~\rm{\textbf{A}}_{1}$. The lower similarity means that the relabelling process enlarges the difference between the annotations and the predictions generated by the proposed method instead of making the annotations adapt to the algorithm's results. All the data in Tab.~\ref{tab:similarity_check} reveals that the annotations of AI-TOD-v2 are essentially distinguished from predictions of NWD-RKA on AI-TOD.

Moreover, we calculate the scale distribution of AI-TOD, the predictions generated by NWD-RKA, and the annotations of AI-TOD-v2, respectively. Results are shown in Fig.~\ref{fig:compare_scale_distribution}. It can be seen that the instance scale distributions of the three sets differ. Especially, the difference is obvious in the scale of $tiny$, NWD-RKA generates about 160,000 more objects than annotations in AI-TOD, where most of he incremented objects are false positive predictions. With manual correction, the number of $tiny$ instances in AI-TOD-v2 eventually comes to 550,933, which is around 40,000 larger than the AI-TOD while 120,000 smaller than the NWD-RKA. The different scale distributions are powerful evidence for the mutual difference between the three sets, while the quantity shrinkage of AI-TOD-v2 compared to predictions of NWD-RKA indicates that the quality of AI-TOD-v2 is elaborately verified in the relabelling process. 

\section{Methodology}
In this section, we first describe the modeling of Normalized Gaussian Wasserstein Distance (NWD) between tiny bounding boxes. We then show the process of our proposed RanKing-based Assignment (RKA) strategy and its combination with NWD. 

\subsection{Normalized Gaussian Wasserstein Distance}

Inspired by the fact that IoU is actually the Jaccard similarity coefficient for computing the similarity of two limited sample sets, we design a better metric for tiny objects based on Wasserstein Distance since it can consistently reflect the distance between distributions even if they have no overlap. Therefore, the new metric has better properties than IoU in measuring similarity between tiny objects. The details are as follows.

For tiny objects, there tend to be some background pixels in their bounding boxes since most~\textit{real objects} are not strict rectangles. In these bounding boxes, foreground pixels and background pixels are concentrated on the center and boundary of the bounding boxes, respectively~\citep{FCOS_2019_ICCV, CenterMap-Net_2020_TGRS}. To better describe the weights of different pixels in bounding boxes, the bounding box can be modeled into two dimensions (2D) Gaussian distribution, where the center pixel of the bounding box has the highest weight, and the importance of the pixel decreases from the center to the boundary. In this paper, we follow the paradigm of taking the center point of the bounding box as the Gaussian distribution mean vector~\citep{CenterMap-Net_2020_TGRS, dense_packed_2019_cvpr, gwd_2021_icml}. Specifically, for horizontal bounding box $R=(cx,cy,w,h)$, where $(cx, cy)$, $w$ and $h$ denote the center coordinates, width and height, respectively. The equation of its inscribed ellipse can be represented as 
\begin{equation}
    \frac{(x-\mu _{x})^2}{\sigma _{x}^{2}}+ \frac{(y-\mu _{y})^2}{\sigma _{y}^{2}}=1,
    \label{eq3.1}
\end{equation}
where $(\mu _{x}, \mu _{y})$ is the center coordinates of the ellipse, $\sigma _{x}$, $\sigma _{y}$ are the lengths of semi-axises along $x$ and $y$ axises. Accordingly, $\mu _{x}=cx, \mu _{y}=cy, \sigma _{x}=\frac{w}{2}, \sigma _{y}=\frac{h}{2}$.

The probability density function of a 2D Gaussian distribution is given by:
\begin{equation}
   f(\mathbf{x}|\boldsymbol{\mu}, \boldsymbol{\Sigma}) = \frac{ \exp (-\frac{1}{2} (\mathbf{x} - \boldsymbol{\mu})^\intercal \boldsymbol{\Sigma}^{-1} (\mathbf{x} - \boldsymbol{\mu}))}{ 2\pi {|\boldsymbol{\Sigma}|}^{\frac{1}{2}}},
   \label{eq3.2}
\end{equation}
where $\mathbf{x}$, $\boldsymbol{\mu}$ and $\mathbf{\Sigma}$ denote the coordinate $(x, y)$, the mean vector and the co-variance matrix of Gaussian distribution. When
\begin{equation}
    (\mathbf{x}-\boldsymbol{\mu})^\intercal \mathbf{\Sigma} ^{-1}(\mathbf{x}-\boldsymbol{\mu})=1,
    \label{eq3.4}
\end{equation}
the ellipse in Eq.~\ref{eq3.1} will be a density contour of the 2D Gaussian distribution. Therefore, the horizontal bounding box $R=(cx,cy,w,h)$ can be modeled into a 2D Gaussian distribution $\mathcal{N}(\boldsymbol{\mu},\boldsymbol{\Sigma})$ with 
\begin{equation}
    \boldsymbol{\mu}=\begin{bmatrix}
        cx \\ cy
    \end{bmatrix}
    ,
    \mathbf{\Sigma }=\begin{bmatrix}
    \frac{w^2}{4} & 0 \\ 
    0 & \frac{h^2}{4}
    \end{bmatrix}.
    \label{eq3.3}
\end{equation}

Furthermore, the similarity between bounding box $A$ and $B$ can be converted to the distribution distance between two Gaussian distributions.

We use the Wasserstein distance which comes from Optimal Transport theory~\citep{optimaltransport_2019} to compute distribution distance. For two 2D Gaussian distributions $\mu_{1}=\mathcal{N}(\boldsymbol{m}_{1}, \boldsymbol{\Sigma}_{1})$ and $\mu_{2}=\mathcal{N}(\boldsymbol{m}_{2}, \boldsymbol{\Sigma}_{2})$, the $2^{\rm nd}$ order Wasserstein distance between $\mu_{1}$ and $\mu_{2}$ can be simplified as:
\begin{equation}
    W_{2}^{2}(\mu_1, \mu_2) =\left\|\mathbf{m}_{1}-\mathbf{m}_{2}\right\|_{2}^{2}+\left\|\boldsymbol{\Sigma}_{1}^{1 / 2}-\boldsymbol{\Sigma}_{2}^{1 / 2}\right\|_{F}^{2}, \\
    \label{sec:simplified_w2}
    \
\end{equation}
where $\left \| \cdot  \right \|_{F}$ is the Frobenius norm. 

For Gaussian distributions $\mathcal{N}_a$ and $\mathcal{N}_b$ which are modeled from bounding boxes $A=({cx}_a, {cy}_a, w_a, h_a)$ and $B=({cx}_b, {cy}_b, w_b, h_b)$, Eq.~\ref{sec:simplified_w2} can be further simplified as:
{\footnotesize
\begin{equation}
    W_{2}^{2}(\mathcal{N}_a, \mathcal{N}_b) = \left\|\left(\left[c x_{a}, c y_{a}, \frac{w_{a}}{2}, \frac{h_{a}}{2}\right]^{\mathrm{T}},\left[c x_{b}, c y_{b}, \frac{w_{b}}{2}. \frac{h_{b}}{2}\right]^{\mathrm{T}}\right)\right\|_{2}^{2}.
\end{equation}
}
However, $W_{2}^{2}(\mathcal{N}_a, \mathcal{N}_b)$ is distance metric with value range $\rm{[0, +\infty]}$, thereby it cannot be directly used as similarity metric for assigning labels. To obtain a value range similar to IoU (\ie, between 0 and 1) and make the metric smooth to location deviations, we heuristically select an exponential nonlinear transformation function to remap the Gaussian Wasserstein distance into another space, thereby normalizing its value range to $\rm{(0, 1]}$. By doing so, we obtain the new metric dubbed Normalized Wasserstein Distance (NWD): 
\begin{equation}
    \textup{NWD}\left(\mathcal{N}_a, \mathcal{N}_b\right)=\exp \left(-\frac{\sqrt{W_{2}^2\left(\mathcal{N}_a, \mathcal{N}_b\right)}}{C}\right),
\end{equation}
where $C$ is a constant, we experimentally observe that its choice is robust in a certain range, details will be shown in Sec.~\ref{sec.ablation}.

\begin{figure*}
    \centering
    \includegraphics[width=0.98\linewidth]{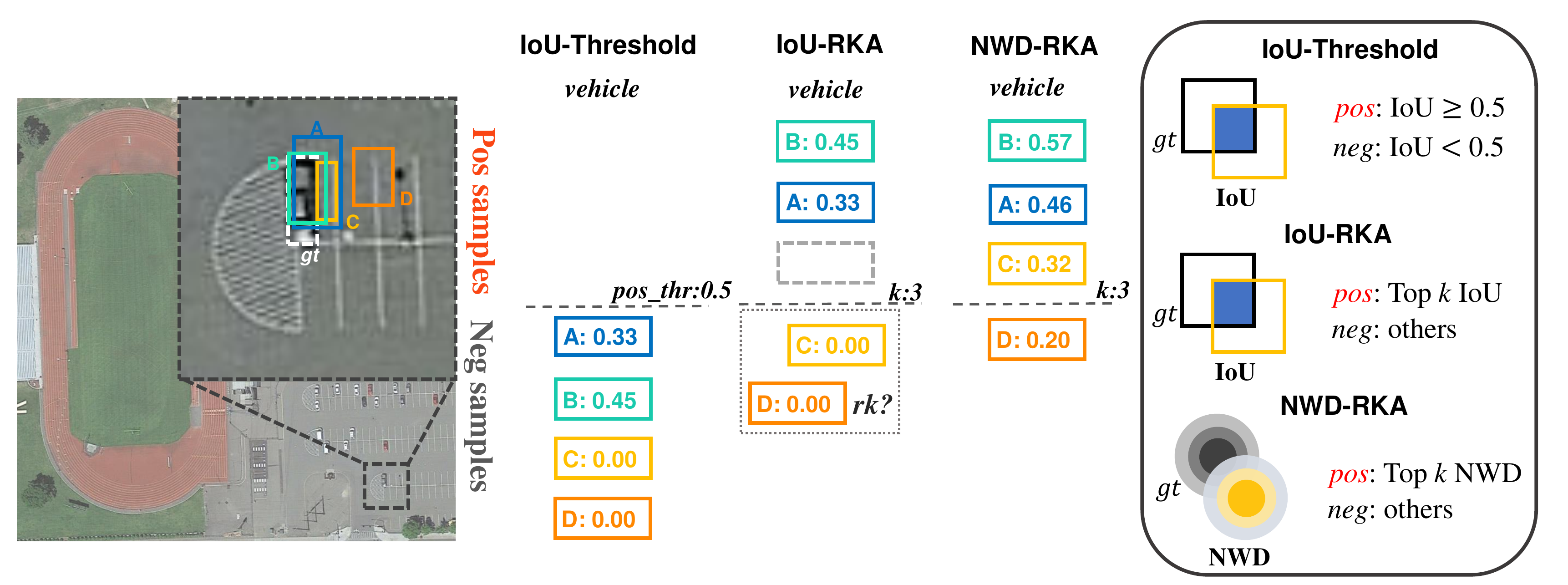}
    \caption{Schematic diagram of NWD-RKA. The white dotted box in the left image denotes the ground truth vehicle. Boxes in other colors denote anchors/proposals. IoU-Threshold represents the baseline assigning strategy in Faster R-CNN, IoU-RKA means combining IoU and our proposed RKA strategy, and the NWD-RKA means combining the proposed NWD and RKA strategy. }
    \label{fig:nwd_rka}
\end{figure*}

\subsection{Ranking-based Assignment}
Herein, we will first revisit the process of traditional threshold-based anchor assignment, which was firstly introduced in Faster R-CNN~\citep{Faster-R-CNN_2015_NIPS}. Our analysis shows that the assignment of anchors for objects of different scales is extremely imbalanced, thus tiny objects are inadequately supervised. To mitigate this problem, we propose a new ranking-based assignment strategy and combine it with NWD to fully leverage the advantage of the new metric on tiny objects.

\textbf{A Revisit of Threshold-based Assignment Strategy.} In Faster R-CNN, anchors of different scales and ratios, are firstly generated. Then binary labels are assigned to the generated anchors to train the classification and regression head of the detector. IoU is used as the metric for determining positive and negative samples. Specifically, the positive label will be assigned to an anchor if it meets one of the two following conditions. The first condition is that the IoU value of a specific anchor and a ground-truth box is the largest of all anchor boxes. The second condition is that the IoU value of a specific anchor and a ground truth box is larger than the positive threshold $\theta_{p}$. Accordingly, the negative label will be assigned to an anchor if its IoU value with all ground truth boxes is smaller than $\theta_{n}$. Anchors that are neither assigned positive labels nor negative labels do not participate in the training process. As shown in Fig.~\ref{fig:num_each_gt}, we compute the mean number of positive training samples assigned to each object in a different size range. It is apparent that under the current IoU-based and threshold-based assignment strategy, the assigner is skewed towards assigning more positive training samples to a certain scale range.

\textbf{Ranking-based Assignment Strategy.} To mitigate the above-mentioned problem and make the network sufficiently supervised by discriminative features of tiny objects, we propose a ranking-based anchor assignment strategy. The schematic diagram is shown in Fig.~\ref{fig:nwd_rka}, key steps of the RanKing-based Assignment (RKA) strategy can be summarized as follows. To begin with, we model the \gt boxes and anchor boxes as 2-D Gaussian distributions and calculate the NWD between each \gt box and anchor box. Then, attaining the score matrix of NWD, we sort each anchor to its NWD score of a particular \gt. Instead of setting a threshold for label assigning, we assign \pos labels to anchors that hold Top $k$ score with a particular \gt, and we assign \textit{neg} labels to remaining anchors. The proposed RKA effectively leverages the advantages of NWD, which will be thoroughly discussed in the following section.

In this paper, we select anchor-based object detectors~\citep{Faster-R-CNN_2015_NIPS, Cascade-R-CNN_2018_CVPR, Focal-Loss_2017_ICCV, DetectoRS_2020_CVPR, ATSS_CVPR_2020} which simultaneously hold the state-of-the-art performance on normal object detection tasks~\citep{DetectoRS_2020_CVPR} and tiny object tasks~\citep{dotd_2021_cvprw} as baseline object detectors. 

Generally, the proposed NWD-RKA strategy can be easily integrated into any anchor-based detector to replace the standard IoU-based assignment strategy, thereby improving its detection performance of tiny objects. It is worth noting that our proposed method will not bring any additional cost during the inference stage.

\section{Analysis}
This section can be divided into two parts. In the first part, we will analyze the characteristics of IoU and NWD. In the second part, we will first summarize three main flaws derived from IoU threshold-based assignment strategy and then reveal how NWD-RKA can handle them simultaneously and improve tiny object detection performance. 

\subsection{Analysis of Metrics}
Compared with IoU, NWD has the following advantages for detecting tiny objects: (1) smoothness to location deviation, (2) scale balance, and (3) the capability of measuring the similarity between non-overlapping or mutually inclusive bounding boxes. As shown in Fig.~\ref{fig::deviation_analysis}, without losing generality, we discuss the change of metric value in the following two scenarios. In the first row of Fig.~\ref{fig::deviation_analysis}, we keep box $A$ and $B$ the same scale and move away $B$ along the diagonal of $A$. It can be observed that IoU is too sensitive to minor location deviation, but the NWD change resulting from location deviation is more smooth. The smoothness to location deviation indicates a possibility of compensating larger number of \pos samples for tiny objects than IoU under the same threshold. Moreover, we can observe that the four curves of NWD completely coincide, which indicates that NWD has the potential to provide a balanced number of training samples to objects of different scales, namely the \textit{scale balance} property. A balanced number of positive training samples is conducive to learning tiny objects since objects of larger scales tend to be assigned with more \pos samples under the current threshold-based label assignment strategy. In the second row of Fig.~\ref{fig::deviation_analysis}, we set the side length of $B$ to half of the side length of $A$ and move away $B$ along the diagonal of $A$. Compared with IoU, the curve of NWD is much more smooth, and it can consistently reflect the similarity between $A$ and $B$ even if $|A\cap B| = A~{\rm or}~B$ and $|A\cap B| = 0$.

\subsection{Analysis of NWD-RKA}
The \textit{pos/neg} imbalance problem is one of the main obstacles to deteriorating detectors' performance, as pointed out in RetinaNet~\citep{Focal-Loss_2017_ICCV}. According to our observation, the ratio of \textit{pos/neg} in AI-TOD-v2 is even much lower than that in MS COCO~\citep{COCO_2014_ECCV} before heuristic sampling~\citep{Faster-R-CNN_2015_NIPS, OHEM_2016_CVPR}, resulting from the IoU's sensitivity to location deviation for tiny objects. It reveals that when detecting tiny objects, the problem of \textit{pos/neg} imbalance is much more severe, leading to the network's lack of supervision information. In this case, the NWD-RKA strategy can significantly mitigate this problem by supplementing more positive samples for the network training. According to our statistics,  using NWD-RKA, the quantity of positive samples is increased to 709,060 (490,240 more than baseline) in one epoch. The incremented positive samples provide promising supervision information for the following classification and regression heads, which can further be deduced from the improvement of AP and AR in Tab.~\ref{tab:baselines}.

\begin{figure}[t]
    \centering
    \subfigure{
    \includegraphics[width=0.95\linewidth]{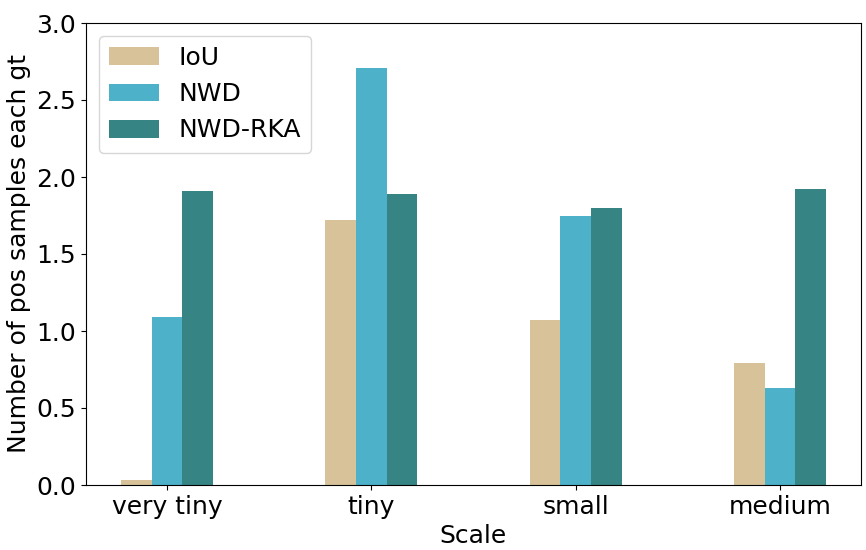}
    }
    \caption{Analysis of number of positive samples assigned to each \gt in different scale ranges. }
    \label{fig:num_each_gt}
    \vspace{-3mm}
\end{figure}

The scale imbalance problem is another central issue hindering scale-invariant object detection~\citep{s3fd_2017_cvpr, imbalance_2020_pami, groupsampling_2019_cvpr}. Concretely, the scale imbalance problem means that the number of positive samples assigned to each \gt is of great variance, resulting from the fact that anchor scales are discrete while object scales are continuous. The scale imbalance problem will lead to the low recall rate of objects far from anchor scales~\citep{s3fd_2017_cvpr}. In tiny object detection, the problem is particularly remarkable. As shown in Fig.~\ref{fig:num_each_gt}, we analyze the number of positive samples assigned to each \gt box under IoU, NWD, and NWD-RKA, respectively. What can be observed is that the imbalance problem is severe in the Faster R-CNN baseline, where objects in the scale range of $very~tiny$ almost fail to be assigned with any positive sample. However, when using the NWD-RKA strategy, the number of positive samples assigned to different scales is notably reconciled, leading to a balanced optimization for different instances.

In order to guarantee that the network sufficiently learns each instance, heuristics are usually deployed to make sure that each \gt is assigned at least one anchor~\citep{Faster-R-CNN_2015_NIPS, RefineDet_2018_CVPR}, and we summarize this as a sample compensation strategy. However, due to the sensitivity of IoU for tiny objects, IoU can easily drop to zero by a minor location deviation, as analyzed in Fig.~\ref{fig::deviation_analysis}. 
Although the Ranking-based Assigner (RKA) is proposed to provide more positive samples for tiny objects based on the ranking of anchor candidates, its advantage cannot be made the most of because the assigner cannot rank the priority of anchors which hold the same IoU value (zero) with a certain \textit{gt}. NWD can easily address this issue since Wasserstein distance can measure the positional relationship when the overlap between distributions is negligible. As shown in Fig.~\ref{fig::deviation_analysis}, even if there is no overlap between the box $A$ and $B$, NWD keeps reflecting their positional relationship (\textit{i.e.}, $\rm{NWD}>0$). In this case, even when there is no overlap (\textit{i.e.}, $\rm{IoU} = 0$) between a specific \textit{gt} and all anchors, the NWD value between them will always be higher than 0, providing a rational metric for ranking. Therefore, at least one appropriate training sample can be guaranteed for this \textit{gt} box under the NWD-RKA strategy.

\begin{table*}
	\centering
	\caption{
	All-class performance of baseline detectors (the first 15 detectors) and our proposed detectors (the last 6 detectors) on the proposed AI-TOD-v2 {\tt test set}. $\rm{AR^{1500}}$ denotes the average recall with a max detection number of 1500. Bold fonts indicate the best performance. In the last row, DetectoRS with our proposed NWD-RKA achieves the state-of-the-art performance of $24.7$ AP.}.
	\begin{tabular}{lccccccccc}  
	\toprule
	Method  & Backbone & AP & $\rm{AP_{0.5}}$ & $\rm{AP_{0.75}}$ & $\rm{AP_{vt}}$ & $\rm{AP_{t}}$ & $\rm{AP_{s}}$ & $\rm{AP_{m}}$ & $\rm{AR^{1500}}$ \\
	\midrule
	\textit{anchor-based two-stage:}      &  	 		& 	&  &  &  &  &  &  &   \\
	  TridentNet~\citep{Trident-Net_2019_ICCV}      & ResNet-50 	 		& 10.1	& 24.5  & 6.7  & 0.1 & 6.3 & 19.8 & 31.9 & 16.4  \\
	  FR w/ ATSS~\citep{ATSS_CVPR_2020}   & ResNet-50-FPN 	 	& 12.8	& 29.6  & 9.2  & 0.0 & 9.2 & 24.9 & 36.6 & 20.5   \\
	  Faster R-CNN~\citep{Faster-R-CNN_2015_NIPS}   & ResNet-50-FPN 	 	& 12.8	& 29.9  & 9.4  & 0.0 & 9.2 & 24.6 & 37.0 & 20.5   \\
	  Faster R-CNN   & ResNet-101-FPN 	 	& 13.1	& 30.7  & 9.2 & 0.0 & 9.7 & 24.6 & 35.5 & 20.6  \\
	  Faster R-CNN   & HRNet-w32 	 	& 14.5	& 32.8  & 10.6 & 0.1 & 11.1 & 27.4 & 37.8 & 22.4 \\
	  Cascade R-CNN~\citep{Cascade-R-CNN_2018_CVPR}  & ResNet-50-FPN 	 	& 15.1	& 34.2  & 11.2 & 0.1 & 11.5 & 26.7 & 38.5 & 22.4 \\
	  DetectoRS~\citep{DetectoRS_2020_CVPR}  & ResNet-50-FPN 	 	& 16.1	& 35.5  & 12.5 & 0.1 & 12.6 & 28.3 & \textbf{40.0} & 23.1 \\
	  DotD~\citep{dotd_2021_cvprw} & ResNet-50-FPN 	& 20.4	& 51.4  & 12.3 & 8.5 & 21.1 & 24.6 & 30.4 & 33.6 \\
    \midrule
    \textit{anchor-based one-stage:}      &  	 		& 	&  &  &  &  &  &  &   \\
	 YOLOv3~\citep{YOLOv3_2017}                  & DarkNet-53 	 	& 4.1	& 14.6  & 0.9  & 1.1 & 4.8 & 7.7  & 8.0 & 11.7   \\
	 RetinaNet~\citep{Focal-Loss_2017_ICCV}      & ResNet-50-FPN 	 	& 8.9	& 24.2  & 4.6  & 2.7 & 8.4 & 13.1  & 20.4 & 25.8   \\
	 SSD-512~\citep{SSD_2016_ECCV}        	   & VGG-16 	 	& 10.7	& 32.5  & 4.0  & 2.0 & 8.7 & 16.8 & 28.0 & 23.5 \\
	\midrule
	\textit{anchor-free:}      &  	 		& 	&  &  &  &  &  &  &   \\
	RepPonits~\citep{RepPoints_2019_ICCV}       & ResNet-50-FPN 	 	& 9.3	& 23.6  & 5.4  & 2.8 & 10.0 & 12.3  & 18.9 & 20.0  \\
	 FoveaBox~\citep{FoveaBox_2020_TIP}        & ResNet-50-FPN 	 	& 11.3	& 28.1  & 7.4  & 1.4 & 8.6 & 17.8  & 32.2 & 21.8  \\
     FCOS~\citep{FCOS_2019_ICCV}                 & ResNet-50-FPN 	 	& 12.0	& 30.2  & 7.3  & 2.2 & 11.1 & 16.6  & 26.9 & 26.0  \\
	 Grid R-CNN~\citep{Grid_RCNN_2019_CVPR}      & ResNet-50-FPN      & 14.3  & 31.1  & 11.0 & 0.1 & 11.0 & 25.7 & 36.7 & 23.5 \\

	\midrule
	\textit{ours:}      &  	 		& 	&  &  &  &  &  &  &   \\
	 RetinaNet w/ NWD-RKA     & ResNet-50-FPN 	 	& 10.5	& 28.5  & 5.2  & 3.5 & 11.0 & 15.2  & 24.2 & 25.7 \\
	 Faster R-CNN w/ NWD-RKA   & ResNet-50-FPN 	 	& 21.4	& 53.2  & 12.5  & 7.7 & 20.7 & 26.8 & 35.2 & 35.7\\
	 Faster R-CNN w/ NWD-RKA   & ResNet-101-FPN 	 	& 20.8	& 52.4  & 12.3  & 8.5 & 20.4 & 24.9 & 35.1 & 34.9 \\
	 Faster R-CNN w/ NWD-RKA   & HRNet-w32 	 	& 22.6	& 55.3  & 14.2  & 8.4 & 21.8 & 27.6 & 36.0 & 36.9\\
	 Cascade R-CNN w/ NWD-RKA  & ResNet-50-FPN 	 	& 22.2	& 52.5  & 15.1 & 7.8 & 21.8 & 28.0 & 37.2 & 35.2\\
	 DetectoRS w/ NWD-RKA  & ResNet-50-FPN 	 	& \textbf{24.7}	& \textbf{57.4}  & \textbf{17.1} & \textbf{9.7} & \textbf{24.2} & \textbf{29.8} & 39.3 &\textbf{37.0} \\
	\bottomrule
	\end{tabular}
	\label{tab:baselines}
\end{table*}

\section{Experiments}
This section focuses on benchmarking some representative object detectors and the effectiveness verification of our proposed method. We first present the experiment settings, including the datasets, parameter settings, and evaluation metrics. Then we provide a large-scale benchmark based on a comprehensive series of detectors. Finally, the effectiveness of our proposed method will be validated by extensive comparative experiments and ablation studies.

\subsection{Experiment Settings}
\textbf{Datasets.}  To evaluate the proposed tiny object detection method, the main experiments are trained on AI-TOD-v2 {\tt trainval set} and validated on AI-TOD-v2 {\tt test set}. To further verify its generalization and robustness, we conducted more experiments on aerial image datasets which contains a considerable number of tiny objects, including: AI-TOD~\citep{AI-TOD_2020_ICPR}, VisDrone2019~\citep{visdrone2019_2019_iccvw}, and DOTA-v2.0~\citep{DOTA2.0_2021_pami}. Note that for VisDrone2019 and DOTA-v2.0, we use their {\tt train set} for training and {\tt validation set} for testing.

\textbf{Parameter Settings.} We conducted all the experiments on a computer with 1 NVIDIA RTX 3090 GPU, and the model training are all based on PyTorch~\citep{PyTorch_2019_NIPS} deep learning framework. We use the ImageNet
pretrained ResNet-50~\citep{ResNet_2016_CVPR} with FPN~\citep{FPN_2017_CVPR} as the backbone, unless specified otherwise. All models are trained using the Stochastic Gradient Descent (SGD) optimizer for 12 epochs with 0.9 momenta, 0.0001 weight decay and 2 batch size. We set the initial learning rate as 0.01 and decay it at epochs 8 and 11. Besides, the batch size of RPN and Faster R-CNN are set to 256 and 512, respectively, and the sampling ratio of positive and negative samples is set to 1/3. The number of proposals generated by RPN is set to 3000. In the inference stage, we use the preset score of 0.05 to filter out background bounding boxes,
and NMS is applied with the IoU threshold of 0.5 to generate the top 3000 confident bounding boxes. The above training and inference parameters are used in all experiments unless specified otherwise.

\textbf{Evaluation Metric.} We use the $\rm{AP}$ (Average Precision) metric to evaluate the performance of the proposed method. 
Specifically, $\rm{AP_{0.5}}$ means the IoU threshold of defining $TP$ is 0.5, $\rm{AP_{0.75}}$ means the IoU threshold of defining $TP$ is 0.75, $\rm{AP}$ means the average value from $\rm{AP_{0.5}}$ to $\rm{AP_{0.95}}$, with IoU interval of 0.05. Note that $\rm{AP_{0.5}}$, $\rm{AP_{0.75}}$ and $\rm{AP}$ take objects of all scales into consideration. Moreover, $\rm{AP}_{vt}$, $\rm{AP}_{t}$, $\rm{AP}_{s}$, and $\rm{AP}_{m}$ are for \textit{very tiny, tiny, small}, and \textit{medium} scale evaluation in AI-TOD~\citep{AI-TOD_2020_ICPR}. $\rm{AR^{1500}}$ denotes average recall with a max detection number of 1500.

\begin{table*}
	\centering
	\caption{Class-wise AP of baseline detectors (the first 15 detectors) and our proposed detectors (the last 6 detectors) on the proposed AI-TOD-v2 {\tt test set}. In this table, we use $\rm{AP}_{0.5:0.95}$ as the evaluation metric of each class, the last row is the $\rm{AP}_{0.5}$ of all classes. The short names for categories are defined as: \textit{airplane (AI), bridge (BR), storage-tank (ST), ship (SH), swimming-pool (SP), vehicle (VE), person (PE) and wind-mill (WM).}}.
	\begin{tabular}{lcccccccccc}  
	\toprule
	Method  & Backbone & AI & BR & ST & SH & SP & VE & PE & WM & $\rm{AP_{0.5}}$ \\
	\midrule
	\textit{anchor-based two-stage:}      &  	 		& 	&  &  &  &  &  &  &   \\
	  TridentNet~\citep{Trident-Net_2019_ICCV}      & ResNet-50 	 		& 19.3	& 0.1  & 17.2  & 16.2 & 12.4 & 12.5 & 3.4 & 0.0 & 24.5  \\
	  FR w/ ATSS~\citep{ATSS_CVPR_2020}   & ResNet-50-FPN 	 	& 24.7	& 5.5  & 20.5  & 19.8 & 12.0 & 15.1 & 4.8 & 0.0 & 29.6  \\
	   Faster R-CNN~\citep{Faster-R-CNN_2015_NIPS}   & ResNet-50-FPN 	 	& 19.7	& 4.8  & 19.0  & 19.9 & 3.7 & 14.4 & 4.8 & 0.0 & 29.9  \\
	   Faster R-CNN   & ResNet-101-FPN 	 	& 25.3	& 8.5  &  19.4 & 19.9 & 12.5 & 14.6  & 4.5 & 0.0 & 30.7 \\
	   Faster R-CNN   & HRNet-w32 	 	& 27.9	& 9.5  & 21.5 & 21.4 & 13.0 & 16.7 & 5.9  & 0.0 & 32.8 \\
	   Cascade R-CNN~\citep{Cascade-R-CNN_2018_CVPR}  & ResNet-50-FPN 	 	& 26.2	& 9.6  & 24.0 & 24.3 & 13.2 & 17.5 & 5.8 & 0.1 & 34.2\\
	   DetectoRS~\citep{DetectoRS_2020_CVPR}  & ResNet-50-FPN 	 	& 28.5	& 11.7  & 23.2 & 26.4 & 14.9 & 17.6 & 6.5 & 0.2 & 35.5\\
	   DotD~\citep{dotd_2021_cvprw} & ResNet-50-FPN 	& 18.7	& 17.5  & 34.7 & 37.0 & 12.4 & 25.4 & 10.3 & 7.4 & 51.4\\
    \midrule
    \textit{anchor-based one-stage:}      &  	 		& 	&  &  &  &  &  &  &   \\
	 YOLOv3~\citep{YOLOv3_2017}                  & DarkNet-53 	 	& 0.3	& 0.5  & 8.5  & 9.4 & 0.0 & 12.7  & 1.4 & 0.0 & 14.6  \\
	 RetinaNet~\citep{Focal-Loss_2017_ICCV}      & ResNet-50-FPN 	 	& 1.3	& 11.8  & 14.3  & 23.6 & 5.8 & 11.4  & 2.3 & 0.5 & 24.2  \\
	 SSD-512~\citep{SSD_2016_ECCV}        	   & VGG-16 	 	& 14.9	& 9.6  & 13.2  & 18.2 & 10.6 & 12.7 & 2.9 & 3.1 & 23.5 \\
	\midrule
	\textit{anchor-free:}      &  	 		& 	&  &  &  &  &  &  &   \\
	 RepPonits~\citep{RepPoints_2019_ICCV}       & ResNet-50-FPN 	 	& 0.0 & 0.1 & 22.5 & 28.8 & 0.2 & 18.3 & 4.1 & 0.0 & 23.6 \\
	 FoveaBox~\citep{FoveaBox_2020_TIP}        & ResNet-50-FPN 	 	& 15.6	& 3.3  & 21.1  & 20.8 & 9.7 & 16.3  & 4.0 & 0.0 & 28.1 \\
     FCOS~\citep{FCOS_2019_ICCV}                 & ResNet-50-FPN 	 	& 7.2 & 13.4 & 20.2 & 26.7 & 8.4 & 16.3 & 3.5 & 0.0 & 30.2\\
	 Grid R-CNN~\citep{Grid_RCNN_2019_CVPR}      & ResNet-50-FPN      & 24.5  & 11.7 & 20.9 & 23.5 & 12.1 & 16.1 & 5.1 & 0.4 & 31.1 \\
	\midrule
	\textit{ours:}      &  	 		& 	&  &  &  &  &  &  &   \\
	 RetinaNet w/ NWD-RKA     & ResNet-50-FPN 	 	& 0.1	& 13.8  & 14.3  & 28.5 & 5.4 & 16.5  & 4.7 & 0.7 & 28.5  \\
	 Faster R-CNN w/ NWD-RKA   & ResNet-50-FPN 	 	& 26.7	& 16.9  & 35.1  & 33.6 & 12.8 & 26.0 & 10.3 & 6.0 & 52.7 \\
	 Faster R-CNN w/ NWD-RKA   & ResNet-101-FPN 	 	& 27.3	& 15.0  & 34.1  & 33.8 & 14.8 & 25.4 & 9.8 & 6.0 & 52.4\\
	 Faster R-CNN w/ NWD-RKA   & HRNet-w32 	 	& 28.4	& 18.4  & 36.8  & 35.0 & 15.5 & \textbf{28.6} & 11.5 & 6.5 & 55.3\\
	 Cascade R-CNN w/ NWD-RKA  & ResNet-50-FPN 	 	& 28.5	& 17.5  & 36.9 & 38.3 & 13.7 & 26.6 & 10.4 & 5.7 & 52.5\\
	 DetectoRS w/ NWD-RKA  & ResNet-50-FPN 	 	& \textbf{32.0}	& \textbf{20.2}  & \textbf{37.8} & \textbf{43.4} & \textbf{16.6} & 27.3 & \textbf{11.6} & \textbf{9.1} & \textbf{57.4} \\
	\bottomrule
	\end{tabular}
	\label{tab:baselines_class}
\end{table*}

\subsection{Benchmark}
We conduct experiments over more than 15 baseline detectors, and the baselines cover both anchor-based (including one-stage and multi-stage) and anchor-free detectors. Note that in this benchmark, we change the number of generated proposals and max number of detections per image from 100 to 3000 since each aerial image usually contains much more objects than each natural image. The results of all class AP and single class AP are respectively listed in Tab.~\ref{tab:baselines} and Tab.~\ref{tab:baselines_class}, where we have the following observations. (1) Generally, when directly applying detectors to AI-TOD-v2, their performance is much worse than that on natural scene datasets such as MS COCO~\citep{COCO_2014_ECCV} and PASCAL VOC~\citep{PASCAL-VOC_2015-IJCV}, in particular, $AP_{vt}$ is close to zero. The extremely limited appearance information and complex background information of tiny objects in aerial images bring enormous challenges to existing detectors. (2) Multi-stage detectors (\eg, Faster R-CNN~\citep{Faster-R-CNN_2015_NIPS}, Cascade R-CNN~\citep{Cascade-R-CNN_2018_CVPR}, and DetectoRS~\citep{DetectoRS_2020_CVPR}) incline to perform better than one-stage anchor-based detectors (\eg, RetinaNet~\citep{Focal-Loss_2017_ICCV}, SSD~\citep{SSD_2016_ECCV}, YOLO~\citep{YOLOv3_2017}) and anchor-free detectors (\eg, FCOS, FoveaBox~\citep{FoveaBox_2020_TIP} and CenterNet~\citep{CenterNet_2019_arXiv}). Benefiting from the progressive optimization mechanism, multi-stage detectors can handle background and foreground class imbalance problems better than anchor-based detectors~\citep{Focal-Loss_2017_ICCV}. Hence, it can be inferred that severe background and foreground class imbalance problems exist for detectors in TOD tasks. (3) The detection performance of some rare categories is much worse than that of common-seen categories, resulting from the imbalanced number of training samples. The imbalance problem also exists in real-life applications and many other aerial datasets~\citep{DIOR_2019_ISPRS, DOTA_2018_CVPR}, requiring the designing of a robust algorithm.

Furthermore, in the real world, label noise problems do exist. As mentioned above, the AI-TOD and AI-TOD-v2 share the same images but contain different annotations. Thus, the AI-TOD and AI-TOD-v2 are a natural pair of \textit{noisy vs. less noisy} datasets. Accurate object detection in the label noise scenario is of great interest both in the computer vision and earth vision community~\citep{noisyanchor_2020-cvpr, learningtolearn_2019_cvpr, rslabelnoise_2017_rs}. In this case, we hope that the existence of AI-TOD will encourage noise-robust object detection methods, which hold promising  improvements on both datasets.

In summary, the proposed large-scale AI-TOD-v2 benchmark systematically poses the problem of tiny object detection in aerial images, and the benchmark is more comprehensive. At the same time, the posed problem is of broad interest in the Earth Vision community and is one of the main obstacles affecting remote sensing interpretation. Moreover, some other interesting problems (\eg, class imbalance, diverse background, label noise) that also exist in other aerial image datasets, will bring many challenges to designing algorithms dedicated to tiny object detection in aerial images. We hope that by introducing the benchmark of detection under the smallest object size, fair comparisons can be made between modified algorithms, thereby promoting the development of the community.

\begin{table}
	\centering
	\caption{Comparison of different metrics for label assignment.}
	\resizebox{\linewidth}{!}{
	\begin{tabular}{cccccccc}  
	\toprule
	Metric & AP & $\rm{AP_{0.5}}$ & $\rm{AP_{0.75}}$ & $\rm{AP_{vt}}$ & $\rm{AP_{t}}$ & $\rm{AP_{s}}$ & $\rm{AP_{m}}$\\
	\midrule
	GWD & 6.0 & 11.9 & 5.3  & 0.0 &  0.3 & 12.7 & 33.5 \\
	GIoU & 10.7 & 23.8 & 8.0 & 0.0 & 5.7 & 23.4 & 35.8 \\
	CIoU & 11.8 & 26.0 & 9.0 & 0.0 & 7.2 & 23.7 & 36.7 \\
	DIoU & 12.1 & 27.0 & 9.4 & 0.0 & 7.6 & 24.6 & \textbf{37.1} \\
	IoU & 12.8 & 29.9 & 9.4 & 0.0  & 9.2 & 24.0 & 37.0 \\
	NWD & \textbf{20.2} & \textbf{49.9} & \textbf{12.5} & \textbf{3.3} & \textbf{19.7} & \textbf{26.9} & 36.4 \\
	\bottomrule
	\end{tabular}}
	\label{tab:assigning}
\end{table}

\begin{table}
    \small
	\centering
	\caption{Ablation of individual effectiveness of NWD and RKA.}
	\resizebox{\linewidth}{!}{
	\begin{tabular}{ccccccccc}  
	\toprule
	NWD & RKA  & AP & $\rm{AP_{0.5}}$ & $\rm{AP_{0.75}}$ & $\rm{AP_{vt}}$ & $\rm{AP_{t}}$ & $\rm{AP_{s}}$ & $\rm{AP_{m}}$ \\
	\midrule
	 & & 12.8 & 29.9 & 9.4 & 0.0 & 9.2 & 24.6 & \textbf{37.0} \\
	 \checkmark & & 20.2 & 49.9 & 12.3 & 3.3 & 19.7 & 26.9 & 36.4 \\
	 & \checkmark & 17.9 & 45.8 & 10.4 & 7.0 & 18.6 & 19.9 & 28.9 \\
	 \checkmark & \checkmark & \textbf{21.4} & \textbf{53.2} & \textbf{12.5} & \textbf{7.7}  & \textbf{20.7} & \textbf{26.8} & 35.2\\
	\bottomrule
	\end{tabular}}
	\label{tab:single}
\end{table}

\subsection{Experimental Results of NWD-RKA}
In this section, we will show that our proposed method significantly eliminates the weaknesses of typical object detectors on tiny objects and  boosts their performance by a large margin. There are four groups of experiments: improvements over baselines, comparison with other assignment metrics, ablation study, and results on different datasets. Note that Faster R-CNN is used as the baseline detector unless otherwise stated. 

\subsubsection{Improvements over baselines}
To verify that our proposed method can be applied into any anchor-based detector and boost tiny object detection performance, we select four detectors for testing, including one-stage anchor-based detector: RetinaNet~\citep{Focal-Loss_2017_ICCV} and two-stage anchor-based detector: Faster R-CNN~\citep{Faster-R-CNN_2015_NIPS}, Cascade R-CNN~\citep{Cascade-R-CNN_2018_CVPR} and DetectoRS~\citep{DetectoRS_2020_CVPR}. Experiment results are shown in Tab.~\ref{tab:baselines} and Tab.~\ref{tab:baselines_class}. It can be seen that NWD-based detectors improve the AP metric of RetinaNet, Faster R-CNN, Cascade R-CNN, and DetectoRS by 1.6, 8.6, 7.1, and 8.6 AP points, respectively. The performance improvement is encouraging and is even more obvious when objects are extremely tiny.

\subsubsection{Comparison with other metrics}
Besides IoU, many refined metrics, such as GIoU~\citep{GIoU_loss_2019_CVPR}, DIoU~\citep{diou_2020_aaai}, CIoU~\citep{diou_2020_aaai}, and GWD~\citep{gwd_2021_icml} are proposed to better evaluate positional relationship between bounding boxes. However, they are originally designed as loss functions and thus may generate sub-optimal results when applied in label assignment. In this paper, to verify the superiority of our proposed method on tiny objects, we respectively re-implement GIoU, DIoU, CIoU, and GWD metrics to replace IoU for label assignment. For a fair comparison, we keep all other parameters the same. 

The comparison results of different metrics of assigning are listed in Tab.~\ref{tab:assigning}. It can be seen that NWD achieves the best performance of 20.2 AP, which suggests that the proposed NWD is more suitable for tiny object measurement and the NWD-based assignment strategy can better leverage the potential of existing detectors on tiny object detection tasks. The poor performance of GWD and IoU-based metrics (\textit{i.e.}, GIoU, DIoU, IoU) mainly results from the fact that they were originally designed as loss functions without a normalized value range. When they are served as the metric for the threshold-based label assignment, a more severe \textit{pos/neg} sample imbalance problem will be introduced, degenerating the TOD performance.

\begin{table}
	\centering
	\caption{The results under different thresholds. In this table, the IoU thresholds are carefully fine-tuned. ($\theta_{p}$, $\theta_{n}$) denotes a pair of positive threshold and negative threshold in RPN.}
	\begin{tabular}{ccccccc}  
	\toprule
	 ($\theta_{p}$, $\theta_{n}$)  & (0.7, 0.3)  & (0.6, 0.2) & (0.5, 0.2) & (0.5, 0.1)  \\
	 \midrule
	 AP  & 12.8 & 16.8 & 16.4 & \textbf{18.6}  \\
	\bottomrule
	\end{tabular}
	\label{tab:threshold}
\end{table}

\vspace{-2mm}

\begin{table}[h]
	\centering
	\caption{Comparison of different anchor size settings under the NWD-RKA strategy.}
	\resizebox{\linewidth}{!}{
	\begin{tabular}{cccccccc}  
	\toprule
	Anchor Scale & AP & $\rm{AP_{0.5}}$ & $\rm{AP_{0.75}}$ & $\rm{AP_{vt}}$ & $\rm{AP_{t}}$ & $\rm{AP_{s}}$ & $\rm{AP_{m}}$\\
	\midrule
	2 & 21.5 & 53.1 & 13.2 & 8.8 & 22.3 & 25.5 & 32.0 \\
	4 & \textbf{22.2} & \textbf{54.0} & \textbf{14.1} & \textbf{8.9} & \textbf{22.3} & \textbf{27.1} & 35.2 \\
	6 & 21.7 & 53.5 & 13.7 & 7.9 & 21.2 & 26.4 & \textbf{35.4}\\
	8 & 21.4 & 53.2 & 12.5 & 7.7  & 20.7 & 26.8 & 35.2\\
	\bottomrule
	\end{tabular}}
	\label{tab:scale}
\end{table}

\begin{table}[h]
    \small
	\centering
	\caption{The AP of different $C$ on different datasets. The $C$ is the normalization factor in NWD. The $k$ in RKA is fixed to 2, 12.7 is the average object size of AI-TOD-v2 {\tt trainval set}. }
	\resizebox{\linewidth}{!}{
	\begin{tabular}{cccccccc}  
	\toprule
	Dataset & $8.0$ & $12.0$ & $12.7$ & $16.0$ &  $20.0$ &  $24.0$ & baseline\\
	\midrule
	AI-TOD-v2 & 21.0 & 21.0  & \textbf{21.4} & 20.8 & 21.0 & 20.7 & 12.8\\
	VisDrone2019 & 23.1 & 23.3 & 23.2 & \textbf{23.5} & 23.1 & 23.4 & 22.3\\
	DOTA-v2.0   & 36.2 & 36.1  & \textbf{36.4} & 36.1 & 36.2 & 36.1 & 35.6 \\
	\bottomrule
	\end{tabular}}
	\label{tab:datasets_c}
\end{table}

\begin{table}[h]
	\centering
	\small
	\caption{The AP performance under different $k$ on different datasets. The $k$ is used for adjusting the number of positive samples in NWD-RKA. Note that $C$ in NWD is fixed to 12.7.}
    \resizebox{\linewidth}{!}{
	\begin{tabular}{cccccccc}  
	\toprule
	Dataset & $1$ & $2$ & $3$ & $4$ & $6$ & $8$ & baseline\\
	\midrule
	AI-TOD-v2 & 20.5 & \textbf{21.4} & 20.9 & 21.0 & 20.9 & 20.7 & 11.1\\
	VisDrone2019 & 23.2 & 23.2 & \textbf{23.4} & 23.0 &  22.9 & 23.0 & 22.3\\
	DOTA-v2.0 & 35.8 & \textbf{36.4} & 36.1 & 36.2 & 35.8 & 36.0 & 35.6 \\
	\bottomrule	\end{tabular}}
	\label{tab:datasets_k}
\end{table}

\begin{table}[h]
	\centering
	\caption{Comparison with different detectors on AI-TOD. Note that we choose ResNet-50 with FPN as backbone. In Tab.~\ref{tab:sota_aitod}, Tab.~\ref{tab:sota_visdrone} and Tab.~\ref{tab:sota_dota}, a detector with * means that the detector is embedded with NWD-RKA.}
	\resizebox{\linewidth}{!}{
	\begin{tabular}{lccccccc}  
	\toprule
	Method & AP & $\rm{AP_{0.5}}$ & $\rm{AP_{0.75}}$ & $\rm{AP_{vt}}$ & $\rm{AP_{t}}$ & $\rm{AP_{s}}$ & $\rm{AP_{m}}$\\
	\midrule
	Faster R-CNN & 11.1 & 26.3 & 7.6 & 0.0  & 7.2 & 23.3 & 33.6\\
	Cascade R-CNN &  13.8  & 30.8  & 10.5  & 0.0  &  10.6 &  25.5 & 36.6  \\
	DetectoRS & 14.8 & 32.8 & 11.4 & 0.0  & 10.8 & 28.3 & \textbf{38.0} \\
	\midrule
	Faster R-CNN* & 19.5 & 49.2 & 11.7 & 8.3  & 19.6 & 24.5 & 31.9\\
	Cascade R-CNN* & 20.5 & 48.7 & 13.8 & 8.1 & 20.6 & 25.6 & 34.0\\
	DetectoRS* & \textbf{23.4} & \textbf{53.5} & \textbf{16.8} & \textbf{8.7}  & \textbf{23.8} & \textbf{28.5} & 36.0\\
	\bottomrule
	\end{tabular}}
	\label{tab:sota_aitod}
\end{table}

\begin{table}[h]
	\centering
	\caption{Comparison with different detectors on VisDrone2019. Note that we choose ResNet-50 with FPN as backbone.}
	\resizebox{\linewidth}{!}{
	\begin{tabular}{lccccccc}  
	\toprule
	Method & AP & $\rm{AP_{0.5}}$ & $\rm{AP_{0.75}}$ & $\rm{AP_{vt}}$ & $\rm{AP_{t}}$ & $\rm{AP_{s}}$ & $\rm{AP_{m}}$\\
	\midrule
	Faster R-CNN & 22.3 & 38.0 & 23.3 & 0.1  & 6.2 & 20.0 & 33.0\\
	Cascade R-CNN &  22.5  & 38.5  & 23.1  & 0.5  &  6.8 &  21.4 & 33.6  \\
	DetectoRS & 25.7 & 41.7 & 27.0 & 0.5  & 7.6 & 23.4 & \textbf{38.1} \\
	\midrule
	Faster R-CNN* & 23.2 & 41.5 & 22.6 & 3.7  & 10.6 & 19.5 & 32.5\\
	Cascade R-CNN* & 23.5 & 40.3 & 23.8 & 3.5 & 10.2 & 19.8 & 33.3\\
	DetectoRS* & \textbf{27.4} & \textbf{46.2} & \textbf{27.7} & \textbf{4.4}  & \textbf{12.6} & \textbf{24.3} & 37.7\\
	\bottomrule
	\end{tabular}}
	\label{tab:sota_visdrone}
\end{table}
\begin{table}[H]
	\centering
	\caption{Comparison with different detectors on DOTA-v2.0. Note that we choose ResNet-50 with FPN as backbone.}
	\resizebox{\linewidth}{!}{
	\begin{tabular}{lccccccc}  
	\toprule
	Method & AP & $\rm{AP_{0.5}}$ & $\rm{AP_{0.75}}$ & $\rm{AP_{vt}}$ & $\rm{AP_{t}}$ & $\rm{AP_{s}}$ & $\rm{AP_{m}}$\\
	\midrule
	Faster R-CNN & 35.6 & 59.5 & 37.2 & 0.0 & 7.1 & 28.9 & 42.1 \\
	Cascade R-CNN & 37.0& 59.5 & 39.6 & 0.0  & 5.9 & 28.4 & 44.0\\
	DetectoRS & 40.8 & 62.6 & 44.4 & 0.0  & 7.0 & 29.9 & 47.8\\
	\midrule
	Faster R-CNN* & 36.4 & 61.5 & 37.6 & 1.5  & 10.4 & 29.4 & 43.2\\
	Cascade R-CNN* & 37.4 & 60.9 & 39.6 & 1.1  & 9.8 & 28.9 & 43.5\\
	DetectoRS* & \textbf{41.9} & \textbf{66.3} & \textbf{44.4} & \textbf{1.9}  & \textbf{10.6} & \textbf{30.3} & \textbf{48.5} \\
	\bottomrule\	\end{tabular}}
	\label{tab:sota_dota}
\end{table}

\subsubsection{Ablation Study}
To verify the effectiveness of the proposed modules individually and to exclude the randomness of hyper-parameters, we design the following five groups of ablation study.   

\textbf{Individual Effectiveness of NWD and RKA.} In this part, we verify the individual effectiveness of NWD and RKA by respectively replacing the IoU and threshold-based anchor assignment in the original network of Faster R-CNN~\citep{Faster-R-CNN_2015_NIPS}. Results are shown in Tab.~\ref{tab:single}. When simply replacing IoU with NWD, the AP can be improved from 12.8 to 20.2. Similarly, when replacing the threshold-based strategy with the ranking-based strategy, the AP is improved from 12.8 to 17.9. Moreover, when combining NWD with RKA, we achieve the best performance of 21.4 AP, implying that the proposed modules can complement each other.

\label{sec.ablation}
\textbf{Different Thresholds of Assigning.} Heuristics are always made to lower the threshold of positive training samples, thereby improving the number of positive training samples for tiny objects~\citep{RefineDet_2018_CVPR}. Indeed, this simple trick can effectively guarantee more positive training samples for tiny objects. However, training samples of low-quality are simultaneously introduced for larger objects. In this section, we carefully fine-tune the threshold of assignment and test as many combinations as possible. Then we compare the results of NWD-RKA with threshold fine-tuning. Results are shown in Tab.~\ref{tab:threshold}, and we can find that the performance of NWD-RKA (21.4 AP) surpasses all different combinations of threshold settings.

\begin{figure*}
    \centering
    \includegraphics[width=\linewidth]{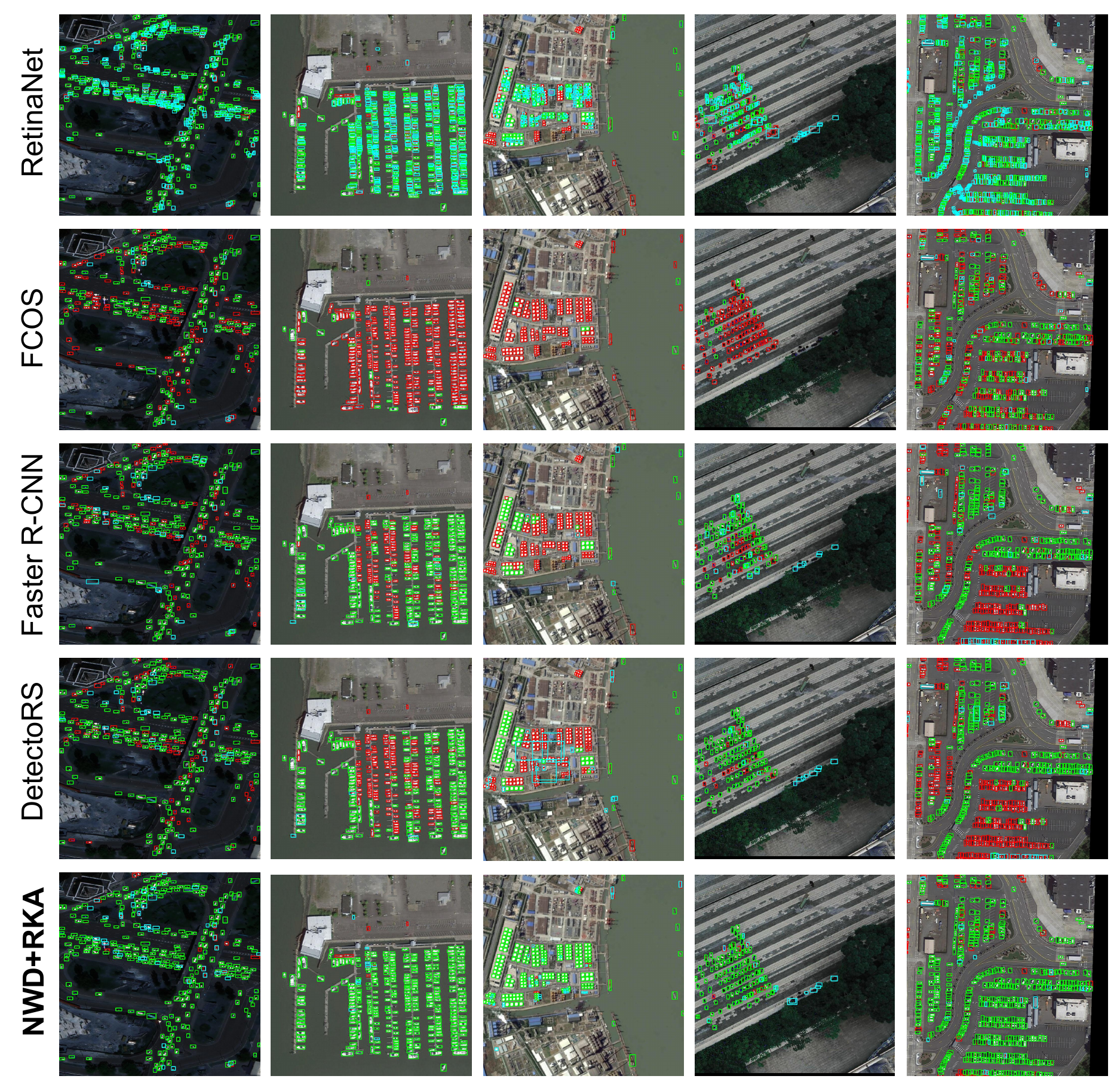}
    \caption{Visualization of detection results using baseline detectors (first four rows) and NWD-RKA based detector (the 5th row) of AI-TOD-v2 dataset. The green, blue, and red boxes denote true positive (TP), false positive (FP), and false negative (FN) predictions, respectively. Note that we apply NWD+RKA into DetectoRS in the 5th row.}
    \label{fig:vis}
\end{figure*}

\textbf{Different Scales of Anchors.} We compare the performance of detectors with NWD-RKA under different anchor scale settings. Results are shown in Tab.~\ref{tab:scale}, since AI-TOD-v2 is a dataset dedicated to TOD, NWD-RKA achieves the best performance of $22.2$ AP with a small anchor scale of $4$. It is worth to be mentioned that our proposed method always keeps a high performance under different anchor size settings, where the performance oscillation is minor. This phenomenon implies that our proposed method is robust to different anchor settings. Thus, the NWD-RKA can be directly applied to detectors without much anchor fine-tuning. Moreover, to guarantee the generality of the proposed method in different scenarios, we retain the default anchor scale setting of $8$ in different tasks.

\textbf{Different Hyper-parameter Value.} There are two hyper-parameters in our design. Herein we show their robustness in a certain range by experiments. When designing NWD, we use a constant $C$ (\ie, the average object size of the training set) to normalize Wasserstein distance, and we have also experimentally set the hyper-parameter to a different constant. The baseline detector is Faster R-CNN embedded with RKA. The results are shown in Tab.~\ref{tab:datasets_c}. We first heuristically set $C$ to the average absolute object size of the {\tt trainval set} of AI-TOD-v2 (12.7) and attain $21.4$ AP. Then, we found that when changing $C$ in a certain range (from 8 to 24), the value of AP waves marginally and is much higher than baseline. It indicates that the 
choice of $C$ is robust in this range. Also, there is a hyper-parameter $k$ in RKA, which is used to adjust the number of positive samples assigned to each instance. We set $k$ to $1$, $2$, $3$, $4$, $6$, and $8$ to test its performance. From the results in Tab.~\ref{tab:datasets_k}, we can see that when setting $k$ to $2$, the best performance can be attained, so $2$ is chosen as the default setting. It can also be observed that the AP waves by a small margin (less than 1) under different $k$. It is worth noting that the choice of parameters (\textit{i.e.},~$C$ and $k$) is also robust in a certain range on different datasets. The results on AI-TOD-v2, VisDrone2019, and DOTA-v2.0 are listed in Tab.~\ref{tab:datasets_c} and Tab.~\ref{tab:datasets_k}, the AP only waves by a minor margin under different parameters. For generality, we recommend fixing $C$ to 12.7 (for simplicity 12) and fixing $k$ to 2 on different datasets.

\textbf{Increasing Feature Resolution.} Tiny objects only contain a few pixels, their feature information will be filtered out after several times of down-sampling. A simple way to handle this problem is to increase the input feature map resolution. In this section, we use Faster R-CNN as the baseline detector and increase the input feature map resolution by replacing ResNet~\citep{ResNet_2016_CVPR} with a stronger backbone (\ie, HRNet~\citep{hrnet_2019_cvpr}). We both conduct the experiment of Faster R-CNN with and without NWD-RKA under the same HRNet backbone. Results are listed in Tab.~\ref{tab:baselines}. Compared with Faster R-CNN with ResNet-50, Faster R-CNN with HRNet has an improvement of 1.7 AP points. However, it is still much lower than Faster R-CNN w/ NWD-RKA with ResNet. Moreover, we add NWD-RKA into Faster R-CNN under the backbone of HRNet and achieve 22.6 AP, which indicates that the proposed NWD-RKA still holds a consistent superiority under a strong backbone.

\subsubsection{Experiments on Other Datasets.}
It is necessary to evaluate the robustness of the proposed method in different datasets. Therefore, we select some other aerial object detection datasets which also contain a considerable number of tiny objects for further verification, including: AI-TOD~\citep{AI-TOD_2020_ICPR}, VisDrone2019~\citep{visdrone2019_2019_iccvw}, and DOTA-v2.0~\citep{DOTA2.0_2021_pami}. 
We embed the proposed NWD-RKA into Faster R-CNN~\citep{Faster-R-CNN_2015_NIPS}, Cascade R-CNN~\citep{Cascade-R-CNN_2018_CVPR} and the state-of-the-art DetectoRS~\citep{DetectoRS_2020_CVPR}. Results are listed in Tab.~\ref{tab:sota_aitod}, Tab.~\ref{tab:sota_visdrone}, and Tab.~\ref{tab:sota_dota}. Significant improvement can be seen on different datasets with all the detectors tested. Concretely, once replacing the IoU threshold-based strategy with NWD-RKA into DetectoRS, 8.6, 1.7, and 1.1 AP boost is attained on AI-TOD, VisDrone2019, and DOTA-v2.0, respectively. Notably, the improvement in the scale range of $tiny$ is remarkable, which further verifies the superiority of NWD-RKA on the TOD task in different scenarios.

\subsection{Visualization}
Some visualization results of baseline detectors and the NWD-RKA-based detectors are shown in Fig.~\ref{fig:vis}. Compared with baseline detectors, a striking improvement can be observed from the detection results of DetectoRS w/ NWD-RKA. Concretely, we have the following observations. The most obvious improvement is that NWD-RKA can greatly eliminate FN. FN is a typical case in baseline detectors when detecting tiny objects, resulting from the lack of supervision information. It indicates that when equipped with NWD-RKA, anchor-based detectors can learn sufficient supervision information from positive tiny object samples. Furthermore, a large number of FP can be seen from the detection results of RetinaNet, which indicates that RetinaNet fails to classify correct predictions from a large number of detection candidates. Surprisingly, NWD-RKA can handle FP detections properly, implying that the assigned \posneg samples are of higher quality. 

\section{Conclusion}
In this paper, we present a novel NWD-RKA method for detecting tiny objects in aerial images while releasing a meticulously optimized dataset AI-TOD-v2 and its corresponding benchmark. The proposed NWD-RKA method is a novel label assignment strategy using normalized Wasserstein distance and ranking-based strategy. Furthermore, the NWD-RKA can be easily embedded into all kinds of anchor-based detectors to boost their performance on tiny object detection. The newly built AI-TOD-v2 holds the smallest average object size in all aerial image datasets. Extensive experiments on four datasets show that our approach can improve the performance of object detectors on tiny objects by a large margin and achieves the state-of-the-art on AI-TOD-v2.

We hope that combining our open-access dataset and promising detection performance will encourage both the earth vision and computer vision communities to consider the challenging problem of tiny object detection in aerial images. Moreover, fair comparisons can be made between customized algorithms, thereby promoting the development of the TOD research.

\section*{Acknowledgements}
We would like to thank Lanxin Zeng, Xiangyuan Wang, Hanqian Si 
and Yan Zhang for the voluntary relabeling work; Haowen Guo and 
Ruixiang Zhang for helpful feedback and discussions. This research was 
partially supported by the National Natural Science Foundation of 
China under Grant 61771351 and the Fundamental Research Funds for 
the Central Universities under Grant 2042022kf1010. The numerical 
calculations were conducted on the supercomputing system in the 
Supercomputing Center, Wuhan University.

\printcredits

\bibliography{Jinwang-Papers, chang}

\end{sloppypar}
\end{document}